\documentclass[10pt,journal,compsoc,x11names]{IEEEtran}
\usepackage[switch]{lineno}
\usepackage{amsmath,amsfonts}
\usepackage{xspace}
\usepackage{enumitem}
\usepackage{amssymb}
\usepackage{booktabs} % For formal tables
\usepackage[hyphens]{url}
\usepackage{ragged2e}
\usepackage{hyperref}
\usepackage{url}
\usepackage{multirow}

\usepackage{algorithmic}
\usepackage{array}
\usepackage[caption=false,font=normalsize,labelfont=sf,textfont=sf]{subfig}
\usepackage{textcomp}
\usepackage{stfloats}
\usepackage{verbatim}
\usepackage{graphicx}

\usepackage{colortbl}
\usepackage{xcolor}
\usepackage{pifont}
\usepackage{resizegather}
\usepackage{algorithm}

\usepackage{svg}
\usepackage{arydshln}
\usepackage{bm}
\usepackage{tikz}
\usepackage[edges]{forest}
\usetikzlibrary{fit}
\definecolor{hidden-draw}{RGB}{205, 44, 36}
\definecolor{hidden-blue}{RGB}{194,232,247}
\definecolor{hidden-orange}{RGB}{243,202,120}
\definecolor{hidden-yellow}{RGB}{242,244,193}
\definecolor{tree-level-1}{RGB}{245,20,85}
\definecolor{tree-level-2}{RGB}{246,86,118}
\definecolor{tree-level-3}{RGB}{248,177,193}
\definecolor{tree-leaf}{RGB}{176,230,198}
\usepackage{lscape}
\usepackage{caption}
\usepackage{subcaption}
\usetikzlibrary{positioning,shapes,shadows,arrows}
\usepackage{bbm}
\usepackage{array}
\usepackage{multirow}
\usepackage{longtable}
\usepackage{rotating, float, caption}
\usepackage{hyperref}

\usepackage{natbib}
\renewcommand{\cite}[1]{\citep{#1}}% define \cite as \citep
\hypersetup{
    colorlinks=true,
    citecolor=blue,
    linkcolor=blue,
    urlcolor=blue
}

\newcolumntype{L}[1]{>{\raggedright\let\newline\\\arraybackslash\hspace{0pt}}m{#1}}
\newcolumntype{C}[1]{>{\centering\let\newline\\\arraybackslash\hspace{0pt}}m{#1}}
\newcolumntype{R}[1]{>{\raggedleft\let\newline\\\arraybackslash\hspace{0pt}}m{#1}}

\newcolumntype{x}[1]{% \raggedleft, \centering or \raggedright
>{\centering\hspace{0pt}}p{#1}}%

\newcommand{\ignore}[1]{}

\usepackage{color, colortbl}
\definecolor{Gray}{gray}{0.9}
\definecolor{LightCyan}{rgb}{0.88,1,1}

\newcommand{\pa}[1]{\noindent \emph{\textbf{#1}}}

\usepackage{amsmath,amsfonts,bm}

\def\eqref#1{equation~\ref{#1}}
\def\1{\bm{1}}

\def\ro{{\textnormal{o}}}

\DeclareMathAlphabet{\mathsfit}{\encodingdefault}{\sfdefault}{m}{sl}
\SetMathAlphabet{\mathsfit}{bold}{\encodingdefault}{\sfdefault}{bx}{n}

\def\gD{{\mathcal{D}}}

\def\gL{{\mathcal{L}}}

\def\gS{{\mathcal{S}}}

\DeclareMathOperator*{\parse}{Parse}

\ifCLASSINFOpdf
\else
\fi

\def\BibTeX{{\rm B\kern-.05em{\sc i\kern-.025em b}\kern-.08em
    T\kern-.1667em\lower.7ex\hbox{E}\kern-.125emX}}
\usepackage{balance}

\begin{document}
\title{A Survey on Knowledge Distillation of Large Language Models}
\author{Xiaohan Xu$^{1}$, Ming Li$^{2}$, Chongyang Tao$^{3}$, Tao Shen$^{4}$, Reynold Cheng$^{1}$, Jinyang Li$^{1}$, \\Can Xu$^{5}$, Dacheng Tao$^{6}$, Tianyi Zhou$^{2}$
\\ \vspace{3mm}
$^1$The University of Hong Kong \quad 
$^2$University of Maryland \quad 
$^3$Microsoft \\
$^4$University of Technology Sydney \quad $^5$Peking University \quad $^6$The University of Sydney\\
{\tt \{shawnxxh,chongyangtao,hishentao\}@gmail.com} \quad  {\tt \{minglii,tianyi\}@umd.edu} \\ \quad {\tt ckcheng@cs.hku.hk}  \quad {\tt jl0725@connect.hku.hk}

}

\markboth{}%
{Shell \MakeLowercase{\textit{et al.}}: Bare Advanced Demo of IEEEtran.cls for IEEE Computer Society Journals}

\IEEEtitleabstractindextext{
\begin{abstract}
\justifying
In the era of Large Language Models (LLMs), Knowledge Distillation (KD) emerges as a pivotal methodology for transferring advanced capabilities from leading proprietary LLMs, such as GPT-4, to their open-source counterparts like LLaMA and Mistral. Additionally, as open-source LLMs flourish, KD plays a crucial role in both compressing these models, and facilitating their self-improvement by employing themselves as teachers. This paper presents a comprehensive survey of KD's role within the realm of LLM, highlighting its critical function in imparting advanced knowledge to smaller models and its utility in model compression and self-improvement. Our survey is meticulously structured around three foundational pillars: \textit{algorithm}, \textit{skill}, and \textit{verticalization} -- providing a comprehensive examination of KD mechanisms, the enhancement of specific cognitive abilities, and their practical implications across diverse fields. Crucially, the survey navigates the interaction between data augmentation (DA) and KD, illustrating how DA emerges as a powerful paradigm within the KD framework to bolster LLMs' performance. By leveraging DA to generate context-rich, skill-specific training data, KD transcends traditional boundaries, enabling open-source models to approximate the contextual adeptness, ethical alignment, and deep semantic insights characteristic of their proprietary counterparts. This work aims to provide an insightful guide for researchers and practitioners, offering a detailed overview of current methodologies in knowledge distillation and proposing future research directions. By bridging the gap between proprietary and open-source LLMs, this survey underscores the potential for more accessible, efficient, and powerful AI solutions. Most importantly, we firmly advocate for compliance with the legal terms that regulate the use of LLMs, ensuring ethical and lawful application of KD of LLMs. An associated Github repository is available at \url{https://github.com/Tebmer/Awesome-Knowledge-Distillation-of-LLMs}.
\end{abstract}

}

\maketitle

\IEEEdisplaynontitleabstractindextext
\IEEEpeerreviewmaketitle

\section{Introduction}

In the evolving landscape of artificial intelligence (AI), proprietary\footnote{For simplicity, we use `proprietary' to represent both versatile yet close-source LLMs like GPT-4 and open-source yet huge LLMs like LLaMA-2-70B, which encapsulate rich knowledge with a large number of parameters.} Large Language Models (LLMs) such as GPT-3.5~\cite{ouyang2022training}, GPT-4~\cite{openai2023gpt4}, Gemini~\cite{team2023gemini} and Claude\footnote{https://www.anthropic.com/claude-in-slack} have emerged as groundbreaking technologies, reshaping our understanding of natural language processing (NLP). These models, characterized by their vast scale and complexity, have unlocked new realms of possibility, from generating human-like text to offering sophisticated problem-solving capabilities. The core significance of these LLMs lies in their emergent abilities~\cite{DBLP:journals/tmlr/WeiTBRZBYBZMCHVLDF22, wei2022chain, xu2024rereading}, a phenomenon where the models display capabilities beyond their explicit training objectives, enabling them to tackle a diverse array of tasks with remarkable proficiency. 
These models excel in understanding and generation, driving applications from creative generation to complex problem-solving~\cite{openai2023gpt4, DBLP:journals/corr/abs-2211-09110}. 
The potential of these models extends far beyond current applications, promising to revolutionize industries, augment human creativity, and redefine our interaction with technology. 

Despite the remarkable capabilities of proprietary LLMs like GPT-4 and Gemini, they are not without their shortcomings, particularly when viewed in light of the advantages offered by open-source models. A significant drawback is their limited accessibility and higher cost~\cite{openai2023gpt4}. These proprietary models often come with substantial usage fees and restricted access, making them less attainable for individuals and smaller organizations. In terms of data privacy and security~\cite{wu2023unveiling}, using these proprietary LLMs frequently entails sending sensitive data to external servers, which raises concerns about data privacy and security. This aspect is especially critical for users handling confidential information. Moreover, the general-purpose design of proprietary LLMs, while powerful, may not always align with the specific needs of niche applications. 
The constraints of accessibility, cost, and adaptability thus present significant challenges in leveraging the full potential of proprietary LLMs.

In contrast to proprietary LLMs, open-source models like LLaMA~\cite{touvron2023llama} and Mistral~\cite{jiang2023mistral} bring several notable advantages. 
One of the primary benefits of open-source models is their accessibility and adaptability. Without the constraints of licensing fees or restrictive usage policies, these models are more readily available to a broader range of users, from individual researchers to smaller organizations. 
This openness fosters a more collaborative and inclusive AI research environment, encouraging innovation and diverse applications. Additionally, the customizable nature of open-source LLMs allows for more tailored solutions, addressing specific needs that generic, large-scale models may not meet.

However, the open-source LLMs also have their own set of drawbacks, primarily stemming from their relatively limited scale and resources compared to their proprietary counterparts. One of the most significant limitations is the smaller model scale, which often results in lower performance on real-world tasks with a bunch of instructions~\cite{DBLP:journals/corr/abs-2306-05685}. These models, with fewer parameters, may struggle to capture the depth and breadth of knowledge embodied in larger models like GPT-4. Additionally, the pre-training investment in these open-source models is typically less substantial. This reduced investment can lead to a narrower range of pre-training data, potentially limiting the models' understanding and handling of diverse or specialized topics~\cite{DBLP:journals/corr/abs-2211-09110, sun2024trustllm}. Moreover, open-source models often undergo fewer fine-tuning steps due to resource constraints. Fine-tuning is crucial for optimizing a model's performance for specific tasks or industries, and the lack thereof can hinder the model's effectiveness in specialized applications. This limitation becomes particularly evident when these models are compared to the highly fine-tuned proprietary LLMs, which are often tailored to excel in a wide array of complex scenarios~\cite{openai2023gpt4}.

% Distillation

\begin{figure}
    \centering
    \includegraphics[width=0.9\linewidth]{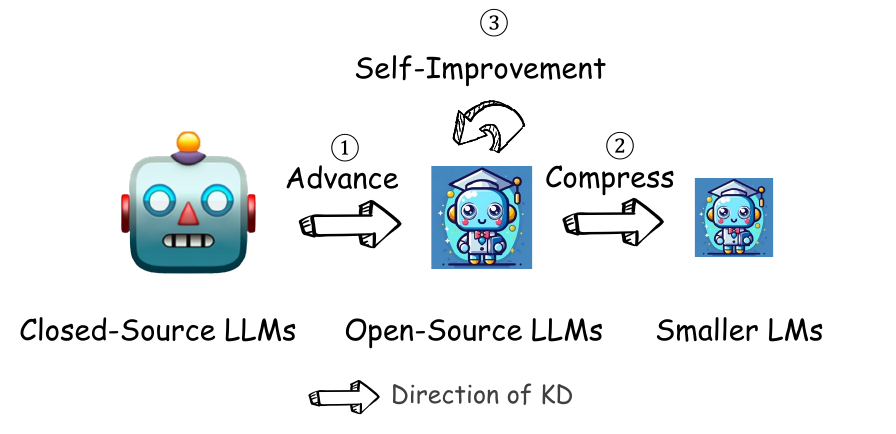}
    \caption{KD plays three key roles in LLMs: 1) Primarily enhancing capabilities, 2) offering traditional compression for efficiency, and 3) an emerging trend of self-improvement via self-generated knowledge.}
    \label{fig:three_roles}
\end{figure}

Primarily, recognizing the disparities between proprietary and open-source LLMs, KD techniques have surged as a means to bridge the performance gap between these models~\cite{gou2021knowledge, gupta2022compression}. Knowledge distillation, in this context, involves leveraging the more advanced capabilities of leading proprietary models like GPT-4 or Gemini as a guiding framework to enhance the competencies of open-source LLMs. This process is akin to transferring the `knowledge' of a highly skilled teacher to a student, wherein the student (e.g., open-source LLM) learns to mimic the performance characteristics of the teacher (e.g., proprietary LLM). Compared to traditional knowledge distillation algorithms~\cite{gou2021knowledge}, data augmentation (DA)~\cite{feng2021survey} has emerged as a prevalent paradigm to achieve knowledge distillation of LLMs, where a small seed of knowledge is used to prompt the LLM to generate more data with respect to a specific skill or domain~\cite{alpaca}. 
Secondly, KD still retains its fundamental role in compressing LLMs, making them more efficient without significant loss in performance. ~\cite{gu2023knowledge, agarwal2023gkd}.
More recently, the strategy of employing open-source LLMs as teachers for their own self-improvement has emerged as a promising approach, enhancing their capabilities significantly~\cite{yuan2024selfrewarding, chen2024selfplay}.  Figure \ref{fig:three_roles} provides an illustration of these three key roles played by KD in the context of LLMs.

A key aspect of the knowledge distillation is the enhancement of skills such as advanced context following (e.g., in-context learning~\cite{huang2022incontext} and instruction following~\cite{alpaca}), improved alignment with user intents (e.g.,  human values/principles~\cite{cui2023ultrafeedback}, and thinking patterns like chain-of-thought (CoT)~\cite{mukherjee2023orca}), and NLP task specialization (e.g., semantic understanding~\cite{ding2023gpt3}, and code generation~\cite{codealpaca}). These skills are crucial for the wide array of applications that LLMs are expected to perform, ranging from casual conversations to complex problem-solving in specialized domains. 
For instance, in vertical domains like healthcare~\cite{wang2023huatuo}, law~\cite{LAWGPT-zh}, or science~\cite{Zhang2024SciGLM}, where accuracy and context-specific knowledge are paramount, knowledge distillation allows open-source models to significantly improve their performance by learning from the proprietary models that have been extensively trained and fine-tuned in these areas.

The benefits of knowledge distillation in the era of LLMs are multifaceted and transformative~\cite{gu2023knowledge}. Through a suite of distillation techniques, the gap between proprietary and open-source models is significantly narrowed~\cite{vicuna2023, xu2023wizardlm} and even filled~\cite{zhao2023survey}. This process not only streamlines computational requirements but also enhances the environmental sustainability of AI operations, as open-source models become more proficient with lesser computational overhead. Furthermore, knowledge distillation fosters a more accessible and equitable AI landscape, where smaller entities and individual researchers gain access to state-of-the-art capabilities, encouraging wider participation and diversity in AI advancements. This democratization of technology leads to more robust, versatile, and accessible AI solutions, catalyzing innovation and growth across various industries and research domains.

The escalating need for a comprehensive survey on the knowledge distillation of LLMs stems from the rapidly evolving landscape of AI~\cite{openai2023gpt4, team2023gemini} and the increasing complexity of these models. As AI continues to penetrate various sectors, the ability to efficiently and effectively distill knowledge from proprietary LLMs to open-source ones becomes not just a technical aspiration but a practical necessity. This need is driven by the growing demand for more accessible, cost-effective, and adaptable AI solutions that can cater to a diverse range of applications and users. A survey in this field is vital for synthesizing the current methodologies, challenges, and breakthroughs in knowledge distillation. It may serve as a beacon for researchers and practitioners alike, guiding them to distill complex AI capabilities into more manageable and accessible forms. Moreover, such a survey can illuminate the path forward, identifying gaps in current techniques and proposing directions for future research.

\begin{figure*}[t]
    \centering
    \includegraphics[width=0.97\linewidth]{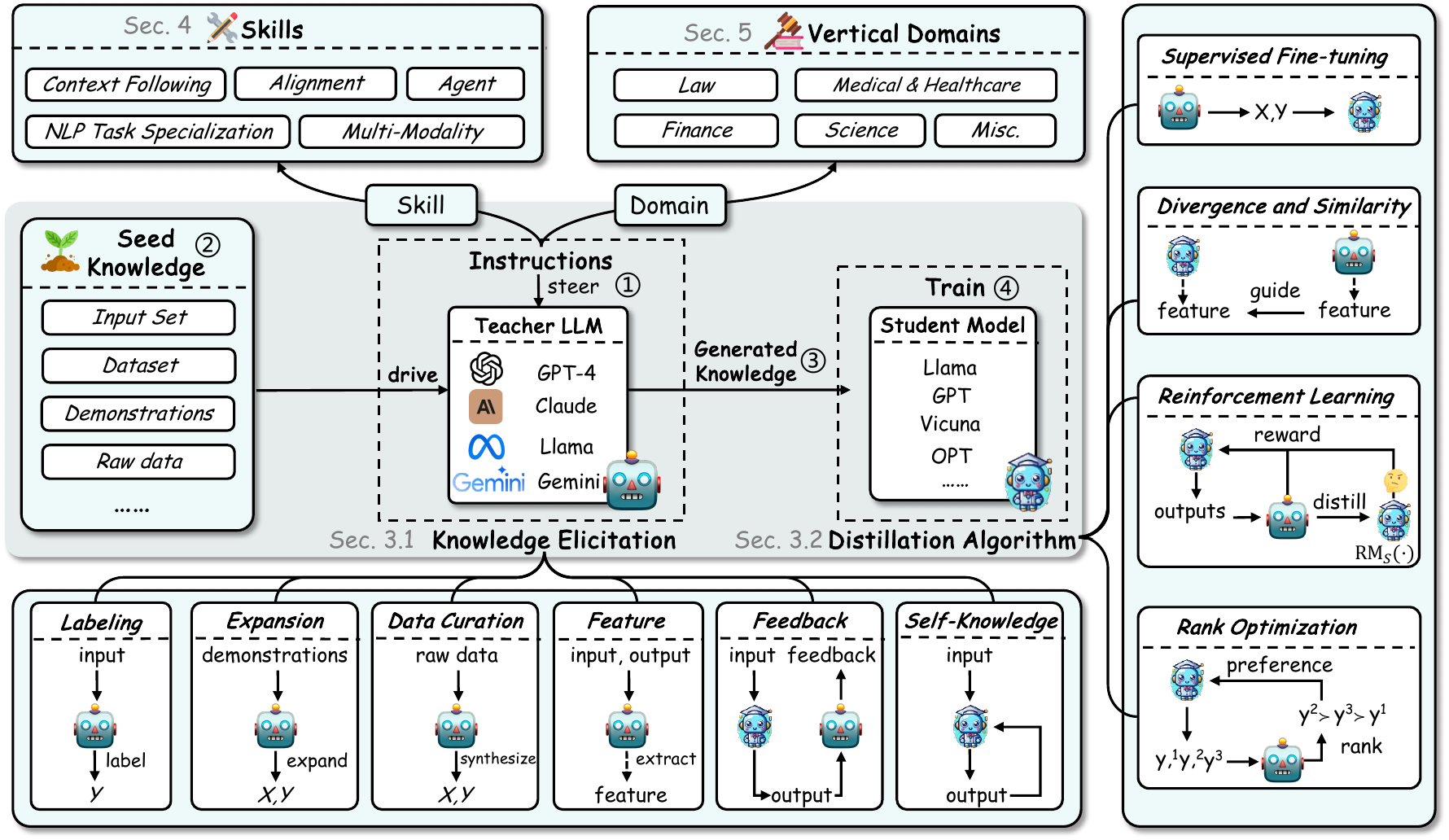}
    \caption{An overview of this survey on knowledge distillation of large language models. Note that `Section' is abbreviated as `Sec.' in this figure. $\text{RM}_S(\cdot)$ denotes the student reward model. \textcircled{\raisebox{-0.9pt}{1}}\textcircled{\raisebox{-0.9pt}{2}}\textcircled{\raisebox{-0.9pt}{3}}\textcircled{\raisebox{-0.9pt}{4}} denote the steps in KD of LLMs.}
    \label{fig:framework}
\end{figure*}

\vspace{2mm}
\pa{Survey Organization.} 
The remainder of this survey is organized into several comprehensive sections, each designed to offer a deep dive into the multifaceted aspects of knowledge distillation within the realm ofLLMs. Following this introduction, \S\ref{sec:overview} provides a foundational overview of knowledge distillation, comparing traditional techniques with those emerging in the era of LLMs and highlighting the role of data augmentation (DA) in this context. \S\ref{sec:kdalgo} delves into the approaches to elicit knowledge from teacher LLMs and core distillation algorithms, examining methods from supervised fine-tuning to more complex strategies involving divergence and similarity, reinforcement learning, and ranking optimization. Then, \S\ref{sec:skill} focuses on skill distillation, exploring how student models can be enhanced to improve context understanding, alignment with user intentions, and performance across a variety of NLP tasks. This includes discussions on natural language understanding (NLU), generation (NLG), information retrieval, recommendation systems, and the evaluation of text generation. In \S\ref{sec:vertical}, we venture into domain-specific vertical distillation, showcasing how knowledge distillation techniques are applied within specialized fields such as law, healthcare, finance, and science, illustrating the practical implications and transformative impact of these approaches. The survey suggests open problems in \S\ref{sec:open}, identifying current challenges and gaps in knowledge distillation research that offer opportunities for future work. Finally, the conclusion and discussion in \S\ref{sec:conclude} synthesize the insights gained, reflecting on the implications for the broader AI and NLP research community and proposing directions for future research. 
Figure~\ref{fig:framework} shows an overview of this survey.

\tikzstyle{my-box}=[
    rectangle,
    rounded corners,
    text opacity=1,
    minimum height=1.5em,
    minimum width=5em,
    inner sep=2pt,
    align=left,
    fill opacity=.5,
]
\tikzstyle{leaf}=[my-box, minimum height=1.5em,
    text=black, align=left,font=\scriptsize,
    inner xsep=2pt,
    inner ysep=4pt,
]
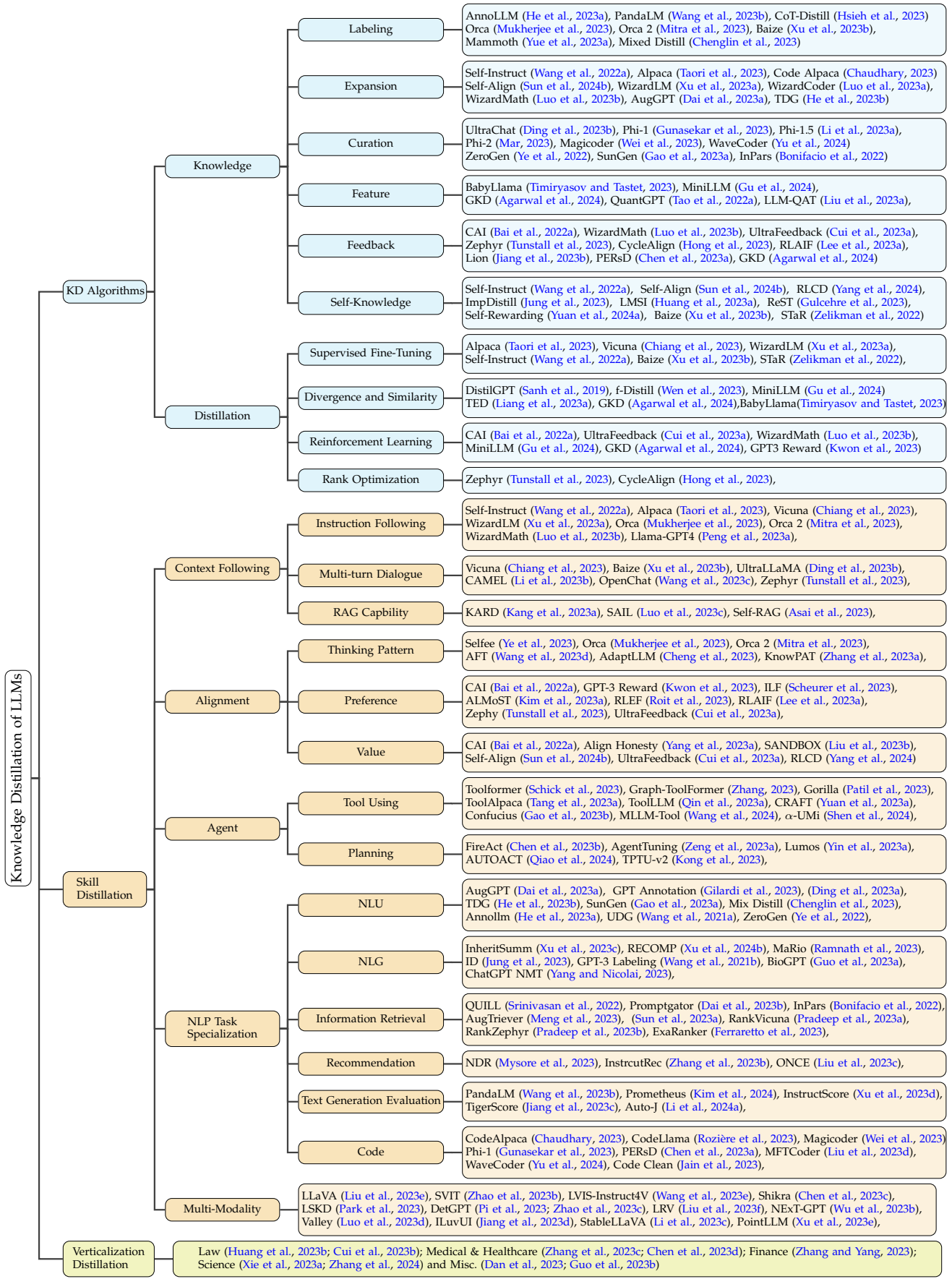
\begin{figure*}[ht!]
    \centering
    \resizebox{0.96\textwidth}{!}{
        \begin{forest}
            forked edges,
            for tree={
                grow=east,
                reversed=true,
                anchor=base west,
                parent anchor=east,
                child anchor=west,
                base=left,
                font=\small,
                rectangle,
                draw=black,
                rounded corners,
                align=left,
                text centered,
                minimum width=4em,
                edge+={darkgray, line width=1pt},
                s sep=3pt,
                inner xsep=2pt,
                inner ysep=3pt,
                ver/.style={rotate=90, child anchor=north, parent anchor=south, anchor=center},
            },
            where level=1{text width=5em,font=\scriptsize,}{},
            where level=2{text width=7em,font=\scriptsize,}{},
            where level=3{text width=9em,font=\scriptsize,}{},
            where level=4{text width=6.1em,font=\scriptsize,}{},
            [
                Knowledge Distillation of LLMs, ver
                [
                    KD Algorithms, for tree={fill=hidden-blue!50}
                    [
                        Knowledge
                        [
                            Labeling
                            [                        
                                AnnoLLM~\cite{he2023annollm}{, }PandaLM~\cite{wang2023pandalm}{, }CoT-Distill~\cite{hsieh2023distilling}\\Orca~\cite{mukherjee2023orca}{, }Orca 2~\cite{mitra2023orca2}{, }Baize~\cite{xu2023baize}{, } \\   Mammoth~\cite{yue2023mammoth}{, }Mixed Distill~\cite{chenglin2023mixed}, leaf, text width=31em
                            ]
                        ]
                        [
                            Expansion
                            [
                                Self-Instruct~\cite{wang2022self}{, }Alpaca~\cite{alpaca}{, }Code Alpaca~\cite{codealpaca}\\Self-Align~\cite{sun2023principledriven}{, }WizardLM~\cite{xu2023wizardlm}{, }WizardCoder~\cite{luo2023wizardcoder}{, }\\WizardMath~\cite{luo2023wizardmath}{, }AugGPT~\cite{dai2023auggpt}{, }TDG~\cite{he-etal-2023-targeted}, leaf, text width=31em
                            ]
                        ]
                        [
                            Curation
                            [
                                UltraChat~\cite{ding2023enhancing}{, }Phi-1~\cite{gunasekar2023textbooks}{, }Phi-1.5~\cite{li2023textbooks1.5}{, }\\Phi-2~\cite{Marah2023phi2}{, }Magicoder~\cite{wei2023magicoder}{, }WaveCoder~\cite{yu2024wavecoder}\\ZeroGen~\cite{ye2022zerogen}{, }SunGen~\cite{DBLP:conf/iclr/GaoPLXY0ZLLK23}{, }InPars~\cite{DBLP:journals/corr/abs-2202-05144}, leaf, text width=31em
                            ]
                        ]
                        [
                            Feature
                            [
                                BabyLlama~\cite{timiryasov2023baby}{, }MiniLLM~\cite{gu2023knowledge}{, }\\GKD~\cite{agarwal2023gkd}{, }QuantGPT~\cite{tao2022compression}{, }LLM-QAT~\cite{liu2023llm}{, }, leaf, text width=31em
                            ]
                        ]
                        [
                            Feedback
                            [
                                CAI~\cite{bai2022constitutional}{, }WizardMath~\cite{luo2023wizardmath}{, }UltraFeedback~\cite{cui2023ultrafeedback}{, } \\ Zephyr~\cite{tunstall2023zephyr}{, }CycleAlign~\cite{hong2023cyclealign}{, }RLAIF~\cite{lee2023rlaif}{, }\\Lion~\cite{jiang2023lion}{, }PERsD~\cite{chen2023personalised}{, }GKD~\cite{agarwal2023gkd},  leaf, text width=31em
                            ]
                        ]
                        [
                            Self-Knowledge
                            [
                                Self-Instruct~\cite{wang2022self}{, } Self-Align~\cite{sun2023principledriven}{, } RLCD~\cite{yang2023rlcd}{, } \\ImpDistill~\cite{jung2023impossible}{, } LMSI~\cite{huang-etal-2023-large}{, } ReST~\cite{gulcehre2023reinforced}{, } \\Self-Rewarding~\cite{yuan2024selfrewarding}{, } Baize~\cite{xu2023baize}{, } STaR~\cite{zelikman2022star}, leaf, text width=31em
                            ]
                        ]
                    ]
                    [
                        Distillation
                        [
                            Supervised Fine-Tuning
                            [
                                Alpaca~\cite{alpaca}{, }Vicuna~\cite{vicuna2023}{, }WizardLM~\cite{xu2023wizardlm}{, }\\Self-Instruct~\cite{wang2022self}{, }Baize~\cite{xu2023baize}{, }STaR~\cite{zelikman2022star}{, }, leaf, text width=31em
                            ]
                        ]
                        [
                            Divergence and Similarity, 
                            [
                                DistilGPT~\cite{sanh2019distilbert}{, }f-Distill~\cite{wen-etal-2023-f}{, }MiniLLM~\cite{gu2023knowledge}\\TED~\cite{liang2023less}{, }GKD~\cite{agarwal2023gkd}{,}BabyLlama\cite{timiryasov2023baby}, leaf, text width=31em
                            ]
                        ]
                        [
                            Reinforcement Learning
                            [
                                CAI~\cite{bai2022constitutional}{, }UltraFeedback~\cite{cui2023ultrafeedback}{, }WizardMath~\cite{luo2023wizardmath}{, }\\MiniLLM~\cite{gu2023knowledge}{, }GKD~\cite{agarwal2023gkd}{, }GPT3 Reward~\cite{kwon2023reward}, leaf, text width=31em
                            ]
                        ]
                        [
                            Rank Optimization
                            [
                                Zephyr~\cite{tunstall2023zephyr}{, }CycleAlign~\cite{hong2023cyclealign}{, }, leaf, text width=31em
                            ]
                        ]
                    ]
                ]
                [
                    Skill \\ Distillation, for tree={fill=hidden-orange!50}
                    [
                        Context Following
                        [
                            Instruction Following 
                            [
                                Self-Instruct~\cite{wang2022self}{, }Alpaca~\cite{alpaca}{, }Vicuna~\cite{vicuna2023}{, }\\WizardLM~\cite{xu2023wizardlm}{, }Orca~\cite{mukherjee2023orca}{, }Orca 2~\cite{mitra2023orca2}{, }\\WizardMath~\cite{luo2023wizardmath}{, }Llama-GPT4~\cite{peng2023instruction}{, }, leaf, text width=31em
                            ]
                        ]
                        [
                            Multi-turn Dialogue
                            [
                                Vicuna~\cite{vicuna2023}{, }Baize~\cite{xu2023baize}{, }UltraLLaMA~\cite{ding2023enhancing}{, }\\CAMEL~\cite{li2023camel}{, }OpenChat~\cite{wang_openchat_2023}{, }Zephyr~\cite{tunstall2023zephyr}{, }, leaf, text width=31em
                            ]
                        ]
                        [
                            RAG Capbility
                            [
                                KARD~\cite{kang2023knowledge}{, }SAIL~\cite{luo2023sail}{, }Self-RAG~\cite{asai2023self}{, }, leaf, text width=31em
                            ]
                        ]
                    ]
                    [
                        Alignment
                        [
                            Thinking Pattern
                            [
                                Selfee~\cite{selfee2023}{, }Orca~\cite{mukherjee2023orca}{, }Orca 2~\cite{mitra2023orca2}{, }\\AFT~\cite{wang2023making}{, }AdaptLLM~\cite{cheng2023adapting}{, }KnowPAT~\cite{zhang2023knowledgeable}{, }, leaf, text width=31em
                            ]
                        ]
                        [
                            Preference
                            [
                                CAI~\cite{bai2022constitutional}{, }GPT-3 Reward~\cite{kwon2023reward}{, }ILF~\cite{scheurer2023training}{, }\\ALMoST~\cite{kim-etal-2023-aligning}{, }RLEF~\cite{roit-etal-2023-factually}{, }RLAIF~\cite{lee2023rlaif}{, }\\Zephy~\cite{tunstall2023zephyr}{, }UltraFeedback~\cite{cui2023ultrafeedback}{, }, leaf, text width=31em
                            ]
                        ]
                        [
                            Value
                            [
                                CAI~\cite{bai2022constitutional}{, }Align Honesty~\cite{yang2023alignment}{, }SANDBOX~\cite{liu2023training}{, }\\Self-Align~\cite{sun2023principledriven}{, }UltraFeedback~\cite{cui2023ultrafeedback}{, }RLCD~\cite{yang2023rlcd},  leaf, text width=31em
                            ]
                        ]
                    ]
                    [
                        Agent
                        [
                            Tool Using
                            [
                            Toolformer~\cite{schick2023toolformer}{, }Graph-ToolFormer~\cite{zhang2023graph}{, }Gorilla~\cite{patil2023gorilla}{, }\\ToolAlpaca~\cite{tang2023toolalpaca}{, }ToolLLM~\cite{qin2023toolllm}{, }CRAFT~\cite{yuan2023craft}{, }\\Confucius~\cite{gao2023confucius}{, }MLLM-Tool~\cite{wang2024mllmtool}{, }$\alpha$-UMi~\cite{shen2024small}{, }, leaf, text width=31em
                            ]
                        ]
                        [
                            Planning
                            [
                                FireAct~\cite{chen2023fireact}{, }AgentTuning~\cite{zeng2023agenttuning}{, }Lumos~\cite{yin2023lumos}{, }\\AUTOACT~\cite{qiao2024autoact}{, }TPTU-v2~\cite{kong2023tptuv2}{, }, leaf, text width=31em
                            ]
                        ]
                    ]
                    [
                        NLP Task \\Specialization
                        [
                            NLU
                            [
                                AugGPT~\cite{dai2023auggpt}{, } GPT Annotation~\cite{Gilardi_2023}{, }\cite{ding2023gpt3}{, }\\TDG~\cite{he-etal-2023-targeted}{, }SunGen~\cite{DBLP:conf/iclr/GaoPLXY0ZLLK23}{, }Mix Distill~\cite{chenglin2023mixed}{, }\\Annollm~\cite{he2023annollm}{, }UDG~\cite{wang2021zerolabel}{, }ZeroGen~\cite{ye2022zerogen}{, }, leaf, text width=31em
                            ]
                        ]
                        [
                            NLG
                            [
                                InheritSumm~\cite{xu-etal-2023-inheritsumm}{, }RECOMP~\cite{xu2023recomp}{, }MaRio~\cite{ramnath2023tailoring}{, }\\ID~\cite{jung2023impossible}{, }GPT-3 Labeling~\cite{wang-etal-2021-want-reduce}{, }BioGPT~\cite{guo2023improving}{, }\\ChatGPT NMT~\cite{yang2023neural}{, }, leaf, text width=31em
                            ]
                        ]
                        [
                            Information Retrieval
                            [
                                QUILL~\cite{srinivasan-etal-2022-quill}{, }Promptgator~\cite{DBLP:conf/iclr/DaiZMLNLBGHC23}{, }InPars~\cite{DBLP:journals/corr/abs-2202-05144}{, }\\AugTriever~\cite{meng2023augtriever}{, }~\cite{sun2023chatgpt}{, }RankVicuna~\cite{pradeep2023rankvicuna}{, }\\RankZephyr~\cite{pradeep2023rankzephyr}{, }ExaRanker~\cite{10.1145/3539618.3592067}{, }, leaf, text width=31em
                            ]
                        ]
                        [
                            Recommendation
                            [
                                NDR~\cite{10.1145/3604915.3608829}{, }InstrcutRec~\cite{zhang2023recommendation}{, }ONCE~\cite{liu2023once}{, }, leaf, text width=31em
                            ]
                        ]
                        [
                            Text Generation Evaluation
                            [
                                PandaLM~\cite{wang2023pandalm}{, }Prometheus~\cite{kim2023prometheus}{, }InstructScore~\cite{xu-etal-2023-instructscore}{, }\\TigerScore~\cite{jiang2023tigerscore}{, }Auto-J~\cite{li2023generative}{, }, leaf, text width=31em
                            ]
                        ]
                        [
                            Code
                            [
                                CodeAlpaca~\cite{codealpaca}{, }CodeLlama~\cite{rozière2023code}{, }Magicoder~\cite{wei2023magicoder}\\Phi-1~\cite{gunasekar2023textbooks}{, }PERsD~\cite{chen2023personalised}{, }MFTCoder~\cite{liu2023mftcoder}{, }\\WaveCoder~\cite{yu2024wavecoder}{, }Code Clean~\cite{DBLP:journals/corr/abs-2311-14904}{, }, leaf, text width=31em
                            ]
                        ]
                    ]
                    [
                        Multi-Modality
                        [
                            LLaVA~\cite{liu2023visual}{, }SVIT~\cite{zhao2023svit}{, }LVIS-Instruct4V~\cite{wang2023believe}{, }Shikra~\cite{chen2023shikra}{, }\\LSKD~\cite{park2023localized}{, }DetGPT~\cite{pi-etal-2023-detgpt,zhao2023chatspot}{, }LRV~\cite{liu2023mitigating}{, }NExT-GPT~\cite{wu2023nextgpt}{, }\\Valley~\cite{luo2023valley}{, }ILuvUI~\cite{jiang2023iluvui}{, }StableLLaVA~\cite{li2023stablellava}{, }PointLLM~\cite{xu2023pointllm}{, }, leaf, text width=41em
                        ]
                    ]
                ]
                [
                    Verticalization \\ Distillation, for tree={fill=hidden-yellow, font=\scriptsize, text width=5.5em}
                    [
                        Law~\cite{huang2023lawyer, cui2023chatlaw}; Medical \& Healthcare~\cite{zhang2023huatuogpt, Chen2023HuatuoGPTII}; Finance~\cite{Zhang2023xuanyuan}; \\Science~\cite{Xie2023DARWIN, Zhang2024SciGLM} and Misc.~\cite{Dan2023EduChat, Guo2023OWL}, text width=49.5em
                    ]
                ]
            ]
        \end{forest}
    }
    \caption{Taxonomy of Knowledge Distillation of Large Language Models. The detailed taxonomy of Verticalization Distillation is shown in Figure~\ref{fig:vertical}.}
    \label{overall}
\end{figure*}

\section{Overview}\label{sec:overview}

\subsection{Comparing Traditional Recipe}

The concept of knowledge distillation in the field of AI and deep learning (DL) refers to the process of transferring knowledge from a large, complex model (teacher) to a smaller, more efficient model (student)~\cite{gou2021knowledge}. This technique is pivotal in mitigating the challenges posed by the computational demands and resource constraints of deploying large-scale models in practical applications.

Historically, knowledge distillation techniques, prior to the era of LLMs, primarily concentrated on transferring knowledge from complex, often cumbersome neural networks to more compact and efficient architectures~\cite{sanh2019distilbert, kim2016sequence}. This process was largely driven by the need to deploy machine learning models in resource-constrained environments, such as mobile devices or edge computing platforms, where the computational power and memory are limited. The focus was predominantly on ad-hoc neural architecture selection and training objectives tailored for single tasks. These earlier methods involved training a smaller student network to mimic the output of a larger teacher network, often through techniques like soft target training, where the student learns from the softened softmax output of the teacher. 
Please refer to the survey~\cite{gou2021knowledge} for more details on general knowledge distillation techniques in AI and DL.

In contrast, the advent of LLMs has revolutionized the knowledge distillation landscape. The current era of knowledge distillation in LLMs shifts the focus from mere architecture compression to knowledge elicitation and transfer~\cite{alpaca, codealpaca, tunstall2023zephyr}. This paradigm change is largely due to the expansive and deep-seated knowledge that LLMs like GPT-4 and Gemini possess. 
And the inaccessible parameters of LLMs make it hard to compress them by using pruning~\cite{han2015deep_compression} or quantization~\cite{liu2023llm} techniques.
Unlike the earlier era, where the goal was to replicate the output behavior of the teacher model or reduce the model size, the current focus in LLM-based knowledge distillation is to elicit the specific knowledge these models have.

The key to this modern approach lies in heuristic and carefully designed prompts, which are used to elicit specific knowledge~\cite{ding2023enhancing} or capabilities~\cite{codealpaca} from the LLMs. These prompts are crafted to tap into the LLM's understanding and capabilities in various domains, ranging from natural language understanding~\cite{he2023annollm} to more complex cognitive tasks like reasoning~\cite{hsieh2023distilling} and problem-solving~\cite{qiao2024autoact}. The use of prompts as a means of knowledge elicitation offers a more flexible and dynamic approach to distillation. It allows for a more targeted extraction of knowledge, focusing on specific skills or domains of interest. This method is particularly effective in harnessing the emergent abilities of LLMs, where the models exhibit capabilities beyond their explicit training objectives.

Furthermore, this era of knowledge distillation also emphasizes the transfer of more abstract qualities such as reasoning patterns~\cite{mitra2023orca2}, preference alignment~\cite{cui2023ultrafeedback}, and value alignment~\cite{sun2023principledriven}.
This is in stark contrast to the earlier focus on output replication~\cite{alpaca}, indicating a shift towards a more holistic and comprehensive transfer of cognitive capabilities. The current techniques involve not just the replication of outputs, but also the emulation of the thought processes~\cite{mitra2023orca2} and decision-making~\cite{asai2023self} patterns of the teacher model. This involves complex strategies like chain-of-thought prompting, where the student model is trained to learn the reasoning process of the teacher, thereby enhancing its problem-solving and decision-making capabilities.

\subsection{Relation to Data Augmentation (DA)}

In the era of LLMs, Data Augmentation (DA)~\cite{wang2022self,  ye2022zerogen} emerges as a critical paradigm integral to the process of knowledge distillation. Unlike traditional DA techniques such as paraphrasing~\cite{gangal2022nareor} or back-translation~\cite{longpre-etal-2019-exploration}, which primarily aim at expanding the training dataset in a somewhat mechanical manner, DA within the context of LLMs focuses on the generation of novel, context-rich training data tailored to specific domains and skills. 

The relationship between DA and KD in LLMs is both symbiotic and foundational. By leveraging a set of seed knowledge, KD employs DA to prompt LLMs to produce explicit data that encapsulates specific skills or domain expertise~\cite{codealpaca, west-etal-2022-symbolic}. This method stands out as a potent mechanism for bridging the knowledge and capability gap between proprietary and open-source models. Through DA, LLMs are prompted to create targeted, high-quality datasets that are not merely larger in volume but are also rich in diversity and specificity. This approach enables the distillation process to be more effective, ensuring that the distilled models not only replicate the teacher model's output behavior but also embody its deep-seated understanding and cognitive strategies.

DA acts as a force multiplier, enabling the distilled models to acquire and refine capabilities that would otherwise require exponentially larger datasets and computational resources. It facilitates a more effective transfer of knowledge, focusing on the qualitative aspects of learning rather than quantitative expansion. This strategic use of DA within KD processes underscores a pivotal shift towards a more efficient, sustainable, and accessible approach to harnessing the power of LLMs. It empowers open-source models with the ability to approximate the contextual adeptness, ethical alignment, and deep semantic insights characteristic of their proprietary counterparts, thereby democratizing access to advanced AI capabilities and fostering innovation across a broader spectrum of applications and users.

\subsection{Survey Scope}

Building on the discussions introduced earlier, this survey aims to comprehensively explore the landscape of knowledge distillation within the context of LLMs, following a meticulously structured taxonomy as in Figure~\ref{overall}. The survey's scope is delineated through three primary facets: KD Algorithms, Skill Distillation, and Verticalization Distillation. 
Each facet encapsulates a range of subtopics and methodologies. It's important to note that KD algorithms provide the technical foundations for skill distillation and verticalization distillation.

\vspace{2mm}
\pa{KD Algorithms.} 
% \textbf{KD Algorithms.}
This segment focuses on the technical foundations and methodologies of knowledge distillation. It includes an in-depth exploration of the processes involved in constructing knowledge from teacher models (e.g., proprietary LLMs) and integrating this knowledge into student models (e.g., open-source LLMs). Under the umbrella of `\emph{knowledge}', we delve into strategies such as labeling~\cite{hsieh2023distilling}, expansion~\cite{alpaca}, curation~\cite{gunasekar2023textbooks}, feature understanding~\cite{agarwal2023gkd}, feedback mechanisms~\cite{tunstall2023zephyr}, and self-knowledge generation~\cite{wang2022self}. This exploration seeks to uncover the various ways in which knowledge can be identified, expanded, and curated for effective distillation. The `\emph{distillation}' subsection examines learning approaches like supervised fine-tuning (SFT)~\cite{wang2022self}, divergence minimization~\cite{agarwal2023gkd}, reinforcement learning techniques~\cite{cui2023ultrafeedback}, and rank optimization strategies~\cite{tunstall2023zephyr}. Together, these techniques demonstrate how KD enables open-source models to obtain knowledge from proprietary ones.

\vspace{2mm}
\pa{Skill Distillation.} This facet examines the specific competencies and capabilities enhanced through KD. It encompasses detailed discussions on context following~\cite{alpaca, luo2023sail}, with subtopics like instruction following and retrieval-augmented generation (RAG) Capability. In the realm of alignment~\cite{mitra2023orca2, tunstall2023zephyr}, the survey investigates thinking patterns, persona/preference modeling, and value alignment. The `agent' category delves into skills such as Tool Using and Planning. NLP task specialization~\cite{dai2023auggpt, jung2023impossible, codealpaca} is scrutinized through lenses like natural language understanding (NLU), natural language generation (NLG), information retrieval, recommendation systems, text generation evaluation, and code generation. Finally, the survey addresses multi-modality~\cite{liu2023visual, zhao2023svit}, exploring how KD enhances LLMs' ability to integrate multiple forms of input.

\vspace{2mm}
\pa{Verticalization Distillation.} This section assesses the application of KD across diverse vertical domains, offering insights into how distilled LLMs can be tailored for specialized fields such as Law~\cite{LAWGPT-zh}, Medical \& Healthcare~\cite{wang2023huatuo}, Finance~\cite{Zhang2023xuanyuan}, Science~\cite{Zhang2024SciGLM}, among others. This exploration not only showcases the practical implications of KD techniques but also highlights their transformative impact on domain-specific AI solutions.

Through these facets, this survey provides a comprehensive analysis of KD in LLMs, guiding researchers and practitioners through methodologies, challenges, and opportunities in this rapidly evolving domain.

\vspace{2mm}
\pa{Declaration.} This survey represents our earnest effort to provide a comprehensive and insightful overview of knowledge distillation techniques applied to LLMs, focusing on algorithms, skill enhancement, and domain-specific applications. Given the vast and rapidly evolving nature of this field, especially with the prevalent practice of eliciting knowledge from training data across academia, we acknowledge that this manuscript may not encompass every pertinent study or development. Nonetheless, it endeavors to introduce the foundational paradigms of knowledge distillation, highlighting key methodologies and their impacts across a range of applications.

\subsection{Distillation Pipeline in LLM Era} 
\label{sec:distillation_pipeline}

\begin{figure}[ht!]
  \centering   
  \includegraphics[width=\linewidth]{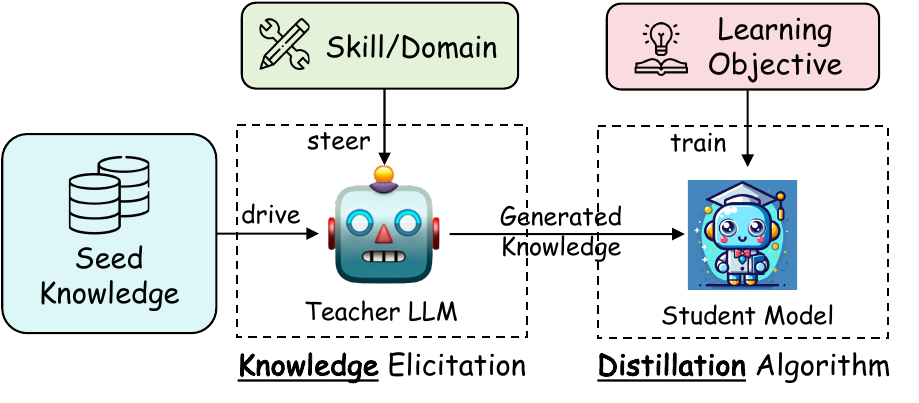}
  \caption{An illustration of a general pipeline to distill knowledge from a large language model to a student model. }
  \label{fig:pipeline}
\end{figure}

The general distillation pipeline of LLMs is a structured and methodical process aimed at transferring knowledge from a sophisticated teacher model to a less complex student model. This pipeline is integral for leveraging the advanced capabilities of models like GPT-4 or Gemini in more accessible and efficient open-source counterparts. The outline of this pipeline can be broadly categorized into four distinct stages, each playing a crucial role in the successful distillation of knowledge. An illustration is shown in Figure~\ref{fig:pipeline}. The detailed pipeline could also be seen in Figure~\ref{fig:framework}.

\textbf{I. Target Skill or Domain Steering Teacher LLM.} The first stage involves directing the teacher LLM towards a specific target skill or domain. This is achieved through carefully crafted instructions or templates that guide the LLM's focus. These instructions are designed to elicit responses that demonstrate the LLM's proficiency in a particular area, be it a specialized domain like healthcare or law, or a skill such as reasoning or language understanding. 

\textbf{II. Seed Knowledge as Input.} Once the target area is defined, the next step is to feed the teacher LLM with seed knowledge. This seed knowledge typically comprises a small dataset or specific data clues relevant to the elicit skill or domain knowledge from the teacher LLM. It acts as a catalyst, prompting the teacher LLM to generate more elaborate and detailed outputs based on this initial information. The seed knowledge is crucial as it provides a foundation upon which the teacher model can build and expand, thereby creating more comprehensive and in-depth knowledge examples.

\textbf{III. Generation of Distillation Knowledge.} In response to the seed knowledge and steering instructions, the teacher LLM generates knowledge examples. These examples are predominantly in the form of question-and-answer (QA) dialogues or narrative explanations, aligning with the natural language processing/understanding capabilities of the LLM. In certain specialized cases, the outputs may also include logits or hidden features, although this is less common due to the complexity and specific requirements of such data forms. The generated knowledge examples constitute the core of the distillation knowledge, encapsulating the advanced understanding and skills of the teacher LLM.

\textbf{IV. Training the Student Model with a Specific Learning Objective.} The final stage involves the utilization of the generated knowledge examples to train the student model. This training is guided by a loss function that aligns with the learning objectives. The loss function quantifies the student model's performance in replicating or adapting the knowledge from the teacher model. By minimizing this loss, the student model learns to emulate the target skills or domain knowledge of the teacher, thereby acquiring similar capabilities. The process involves iteratively adjusting the student model's parameters to reduce the discrepancy between its outputs and those of the teacher model, ensuring the effective transfer of knowledge.

In essential, the above four stages can be abstracted as two formulations. The first formulation represents the process of eliciting knowledge:

\begin{equation}\label{eq:data}
     \mathcal{D}^{\text{(kd)}}_I = \{\parse(o, s) | o \sim p_T(\ro |I \oplus s), \forall s \sim \mathcal{S}\},
\end{equation}    
where $\oplus$ denotes fusing two pieces of text, $I$ denotes an instruction or a template for a task, skill, or domain to steer the LLM and elicit knowledge,
$s \sim \gS$ denotes an example of the seed knowledge, upon which the LLM can explore to generate novel knowledge, 
$\parse(o, s)$ stands for to parse the distillation example ( e.g., $(x, y)$) from the teacher LLM's output $o$ (plus the input $s$ in some cases), and $p_{T}$ represents the teacher LLM with parameters $\theta_T$.
Given the datasets $\gD^{\text{(kd)}}_I$ built for distillation, we then define a learning objective as
\begin{align}\label{eq:objective}
    \mathcal{L} = \sum\nolimits_I \gL_I(\mathcal{D}^{\text{(kd)}}_I;\theta_S),
\end{align}
where $\sum\nolimits_I$ denotes there could be multiple tasks or skills being distilled into one student model, $\gL_I(\cdot;\cdot)$ stands for a specific learning objective, and $\theta_S$ parameterizes the student model. 

Following our exploration of the distillation pipeline and the foundational concepts underlying knowledge distillation in the LLM era, we now turn our focus to the specific algorithms that have gained prominence in this era.

\begin{figure*}[h]
    \centering
    \includegraphics[width=\linewidth]{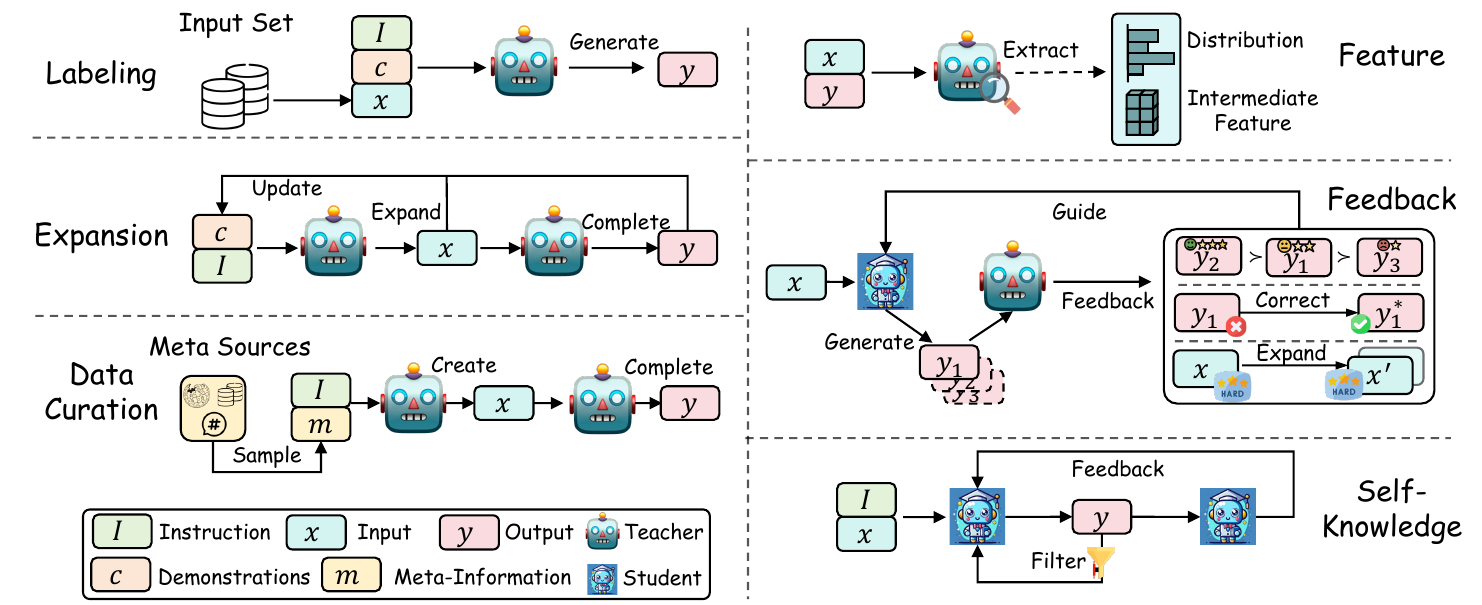}
    \caption[Caption Text]{An illustration of different knowledge elicitation methods from teacher LLMs. \textit{Labeling}: The teacher generates the output from the input; \textit{Expansion}: The teacher generates samples similar to the given demonstrations through in-context learning; \textit{Data Curation}: The teacher synthesizes data according to meta-information, such as a topic or an entity; \textit{Feature}: Feed the data into the teacher and extract its internal knowledge, such as logits and features; \textit{Feedback}: The teacher provides feedback on the student's generations, such as preferences, corrections, expansions of challenging samples, etc; \textit{Self-Knowledge}: The student first generates outputs, which is then filtered for high quality or evaluated by the student itself.}
    \label{fig:knowledge}
\end{figure*}

\section{Knowledge Distillation Algorithms}\label{sec:kdalgo}
\label{sec:kd}

This section navigates through the process of knowledge distillation. According to Section \ref{sec:distillation_pipeline}, it is categorized into two principal steps: `Knowledge,' focusing on eliciting knowledge from teacher LLMs (Eq.\ref{eq:data}), and `Distillation,' centered on injecting this knowledge into student models (Eq.\ref{eq:objective}). 
We will elaborate on these two processes in the subsequent sections.

\subsection{Knowledge}
This section focuses on the approaches to elicit knowledge from teacher LLMs. According to the manners to acquire knowledge, we divided them into \textit{Labeling}, \textit{Expansion}, \textit{Data Curation}, \textit{Feature}, \textit{Feedback}, and \textit{Self-Knowledge}. Figure \ref{fig:knowledge} shows an illustration of these knowledge elicitation methods.

\subsubsection{Labeling} 

Labeling knowledge refers to using a teacher LLM to label the output \(y\) for a given input \(x\) as the seed knowledge, according to the instruction \(I\) or demonstrations \(c\), where \(c = (x_1, y_1), \ldots, (x_n, y_n)\). This method of eliciting knowledge from teacher LLMs is straightforward yet effective and has been widely applied across various tasks and applications. It requires only the collection of an input dataset and feeding it into LLMs to obtain the desired generations. Moreover, the generation of \(y\) is controllable through the predefined \(I\) and \(c\). This process can be formulated as follows:
\begin{equation}
    \mathcal{D}^{\text{(lab)}} = \{x, y|x \sim \mathcal{X}, y \sim p_{T}(y |I \oplus c \oplus x)\}.
\end{equation}

Input $x$ could be sourced from existing NLP task datasets, which serve as typical reservoirs for distillation efforts.
Numerous works have sought to harness the capabilities of powerful LLMs as teachers for annotating dataset samples across a range of tasks.
For instance, efforts in natural language understanding involve using LLMs to categorize text~\cite{Gilardi_2023, ding2023gpt3, he2023annollm}, while in natural language generation, LLMs assist in generating sequences for outputs~\cite{hsieh2023distilling, jung2023impossible, wang-etal-2021-want-reduce}. Text generation evaluation tasks leverage LLMs to label evaluated results~\cite{li2024leveraging, wang2023pandalm}, and reasoning tasks utilize LLMs for labeling Chains of Thought (CoT) explanations~\cite{hsieh2023distilling, li2022explanations, ho2023large, magister-etal-2023-teaching, fu2023specializing, ramnath2023tailoring, li-etal-2023-symbolic, liu2023minds}, among others.
Rather than concentrating on specific tasks, many current works focus on labeling outputs based on instructions, thereby teaching student models to solve tasks in a more flexible way by following instructions.
Collections of various NLP tasks, complemented by instructional templates, serve as valuable input sources for $x$. For instance, FLAN-v2 collections~\cite{longpre2023flan} offers extensive publicly available sets of tasks with instructions, which are labeled with responses generated by teacher LLMs in Orca~\cite{mukherjee2023orca, mitra2023orca2}. 
The instructions from these NLP tasks are built from predefined templates, which lack diversity and may have gaps between human's natural query. 
The real conversations between humans and chat models provide large-scale data with real queries and generations labeled by powerful LLMs, like ShareGPT. 
Additionally, \citet{xu2023baize} and \citet{anand2023gpt4all} label the real questions sampled from forums like Quora and Stack Overflow.

Moreover, the process of labeling could be guided by instructions $I$ or demonstrations $c$. 
A commonly used instruction type for guiding labeling is chain-of-thought (CoT) prompt~\cite{hsieh2023distilling, fu2023specializing, magister-etal-2023-teaching}.
\citet{mukherjee2023orca} add multiple system messages (e.g. ``You must generate a detailed and long answer.'' or ``explain like I’m five, think step-by-step'') to elicit rich signals. \citet{yue2023mammoth} and \citet{chenglin2023mixed} label a hybrid of knowledge of chain-of-thought (CoT) and program-of-thought (PoT) rationales.
\citet{xu2023baize} propose a self-chat technique that two teacher LLMs simulate the real conversational to generate multi-turn dialogues for a question from Quora and Stack Overflow.

\subsubsection{Expansion}

While the labeling approach is simple and effective, it faces certain limitations. Primarily, it is constrained by the scale and variety of the input data. In real-world applications, especially those involving user conversations, there are also concerns regarding the privacy of the data involved. To address these limitations, various expansion methods have been proposed~\cite{wang2022self, alpaca, codealpaca, si2023empirical, ji2023exploring, luo2023wizardmath, luo2023wizardcoder, wu2023laminilm, sun2023principledriven, xu2023wizardlm, guo2023instruction, rozière2023code, west-etal-2022-symbolic}. These methods take the demonstrations as seed knowledge and aim to expand a large scale and various data by in-context learning.

A key characteristic of these expansion methods is the utilization of the in-context learning ability of LLMs to generate data similar to the provided demonstrations $c$. Unlike in the labeling approach, where the input \( x \) is sampled from the existing dataset, in the expansion approach, both \( x \) and \( y \) are generated by teacher LLMs. This process can be formulated as follows:
\begin{equation}
\begin{aligned}
    \mathcal{D}^{\text{(exp)}} = \{(x, y)| x \sim p_T(x |I \oplus c), y \sim p_T(y |I \oplus x)\}.
    \end{aligned}
\end{equation}
In this formulation, \( x \) and \( y \) represent the new input-output pairs generated by the teacher LLM. The input \( x \) is generated based on a set of input-output demonstrations \( c \). The output \( y \) is then generated in response to the new input \( x \) under the guidance of an instruction \( I \). Note that the demonstrations could be predefined or dynamically updated by adding the newly generated samples.

Expansion techniques have been widely utilized to extract extensive instruction-following knowledge from teacher LLMs.
\citet{wang2022self} first introduces an iterative bootstrapping method, Self-Instruct, to utilize LLMs to generate a wide array of instructions based on several demonstrations sampled from 175 manually-written instructions. The newly generated instructions are then added back to the initial pool, benefiting subsequent expansion iterations.
Subsequently, \citet{alpaca} applies this expansion method to a more powerful teacher LLM, text-davinci-003, to distill 52K high-quality data.
To improve the diversity and coverage during expansion, \citet{wu2023laminilm} and \cite{sun2023principledriven} prompt the teacher LLM to generate instructions corresponding to some specific topics. 
\citet{xu2023wizardlm} propose an Evol-Instruct method to expand the instructions from two dimensions: difficulty (e.g. rewriting the question to be more complex) and diversity (e.g. generating more long-tailed instructions). This Evol-Instruct method is domain-agnostic and has been used to expand the distillation of coding~\cite{luo2023wizardcoder} and math~\cite{luo2023wizardmath}.
Additionally, expansion methods can significantly augment NLP task datasets with similar samples, thereby enhancing task performance. 
For instance, AugGPT~\cite{dai2023auggpt} leverages a teacher LLM to rephrase each sentence in the training samples into multiple conceptually similar, but semantically varied, samples to improve classification performance. 
Similarly, TDG~\cite{he-etal-2023-targeted} proposes the Targeted Data Generation (TDG) framework, which automatically identifies challenging subgroups within data and generates new samples for these subgroups using LLMs through in-context learning.

In summary, the expansion method leverages the in-context learning strengths of LLMs to produce more varied and extensive datasets with both inputs and outputs.
However, the quality and diversity of the generated data are heavily reliant on the teacher LLMs and the initial seed demonstrations. This dependence can lead to a dataset with inherent bias from LLMs~\cite{yu2023large, wei2023magicoder} and a homogeneity issue where the generations may be prone to similarity ultimately, 
limiting the diversity this method seeks to achieve~\cite{ding2023enhancing}.
Moreover, the expansion process may inadvertently amplify any biases present in the seed data.

\subsubsection{Data Curation}

The pursuit of high-quality and scalable data generation in knowledge distillation from LLMs has led to the emergence of the Data Curation approach. This method arises in response to the limitations observed in both the Labeling and Expansion approaches. These methods often yield data of variable quality and face constraints in quantity. In Labeling, the seed knowledge is sourced from task datasets, leading to potential noise and dirty data. Meanwhile, in Expansion, the input \( x \) is derived from seed demonstrations, which can result in homogeneous data when generated in large quantities. To overcome these challenges, the Data Curation method curates high-quality or large-scale data by extensive meta-information as seed knowledge~\cite{ding2023enhancing, gunasekar2023textbooks,li2023textbooks1.5,Marah2023phi2, liu2023mftcoder, wei2023magicoder, yu2024wavecoder, ye2022zerogen, DBLP:conf/iclr/GaoPLXY0ZLLK23, yang2023neural}.

A distinct feature of Data Curation is its approach to synthesize data from scratch. 
Numerous diverse meta-information, such as topics or knowledge points, could be incorporated into this process to generate controllable \( x \) and \( y \).
Thus, this process can be meticulously controlled to yield datasets that are not only large in scale but also of high quality. The formulation for Data Curation can be represented as:
\begin{equation}
\begin{aligned}
    \mathcal{D}^{\text{(cur)}} = \{(x, y)| x \sim p_T(x |I \oplus m ), y \sim p_T(y |I \oplus x)\}.
\end{aligned}    
\end{equation}
In this formulation, \( m \) represents the diverse meta-information used to guide the  synthesis of \( x \), and \( I \) is the instruction guiding teacher LLMs to generate \( x \) or \( y \).

Different studies primarily vary in their source and method of leveraging meta-information. 
UltraChat~\cite{ding2023enhancing} effectively demonstrates the process of curating both high-quality and diverse data by distilled knowledge. They collect extensive meta-information across three domains: \textit{Questions about the World, Creation and Generation}, and \textit{Assistance on Existing Materials}. For example, under \textit{Questions about the World}, they explore 30 meta-topics like "Technology" and "Food and Drink." the teacher LLMs then use this meta-information to distill a broad array of instructions and conversations, achieving a substantial scale of 1.5 million instances. UltraChat stands out with its lexical and topical diversity. The UltraLLaMA model, fine-tuned on this data, consistently surpasses other open-source models.
Another notable series, \textbf{phi}~\cite{gunasekar2023textbooks,li2023textbooks1.5,Marah2023phi2}, focuses on distilling smaller, high-quality datasets akin to "textbooks." \textbf{Phi-1}\cite{gunasekar2023textbooks} experiments with synthesizing "textbook quality" data in the coding domain. Their approach involves distilling clear, self-contained, instructive, and balanced content from LLMs, guided by random topics or function names to enhance diversity. The distilled data is a synthesis of 1 billion tokens of Python textbooks, complete with natural language explanations and code snippets, as well as 180 million tokens of Python exercises with solutions. Remarkably, the \textbf{phi-1} model, despite its smaller size, outperforms nearly all open-source models on coding benchmarks like HumanEval and MBPP while being 10 times smaller in model size and 100 times smaller in dataset size.
MFTCoder~\cite{liu2023mftcoder} utilizes hundreds of Python knowledge points as meta-information to create a CodeExercise Dataset. In contrast, Magicoder~\cite{wei2023magicoder} and WaveCoder~\cite{yu2024wavecoder} get raw code collections from open-source code datasets, using this as meta-information for generating instructional data. In the context of NLU tasks, certain studies~\cite{ye2022zerogen, DBLP:conf/iclr/GaoPLXY0ZLLK23, wang2021zerolabel} explore the use of labels as meta-information to synthesize corresponding samples for data augmentation. 
Similarly, in information retrieval tasks, there are efforts to utilize documents as meta-information for generating potential queries, thereby constructing large-scale retrieval pairs~\cite{DBLP:journals/corr/abs-2202-05144, meng2023augtriever}.

In conclusion, Data Curation through teacher LLMs has emerged as a promising technique for synthesizing datasets that are not only high-quality and diverse but also large in scale. 
The success of models like \textbf{phi-1} in specialized domains underscores the efficacy of this method. 
The ability to create synthetic datasets will become a crucial technical skill and a key area of focus in AI~\cite{li2023textbooks1.5}.

\subsubsection{Feature}

The previously discussed knowledge elicitation methods are typically applied to powerful black-box models, which are expensive and somewhat unreproducible due to calling API. In contrast, white-box distillation offers a more transparent and accessible approach for researchers. It involves leveraging the output distributions, intermediate features, or activations from teacher LLMs, which we collectively refer to as \textit{Feature} knowledge.
White-box KD approaches have predominantly been studied for smaller encoder-based LMs, typically those with fewer than 1 billion parameters (cf. \citet{gou2021knowledge} for detail). However, recent research has begun to explore white-box distillation in the context of generative LLMs~\cite{timiryasov2023baby, liang2023less, gu2023knowledge, agarwal2023gkd, liu2023llm, wen-etal-2023-f, wan2024knowledge, zhao2023towards, qin2023improving, boizard2024towards, zhong2024revisiting}. 

The typical method for acquiring this feature knowledge involves teacher LLMs annotating the output sequence \( y \) with its internal representations. These annotations are then distilled into the student model using methods such as Kullback-Leibler Divergence (KLD).
The process of eliciting feature knowledge can be formulated as follows:
\begin{equation}
\begin{aligned}
    \mathcal{D}^{\text{(feat)}} = \{ (x, y, \phi_{\text{feat}}(x, y; \theta_T)) \mid x \sim \mathcal{X}, y \sim \mathcal{Y} \}.
\end{aligned}
\end{equation}
In this formulation, $\mathcal{Y}$ is the output set, which can be generated by teacher LLMs, the student model, or directly sourced from the dataset.
\( \phi_\text{feat}(\cdot; \theta_T) \) represents the operation of extracting feature knowledge (such as output distribution) from the teacher LLM.

The most straightforward method to elicit feature knowledge of teacher is to label a fixed dataset of sequences with token-level probability distributions~\cite{sanh2019distilbert, wen-etal-2023-f}.
To leverage the rich semantic and syntactic knowledge in intermediate layers of the teacher model, TED ~\cite{liang2023less} designs task-aware layer-wise distillation. They align the student's hidden representations with those of the teacher at each layer, selectively extracting knowledge pertinent to the target task.
\citet{gu2023knowledge} and \citet{agarwal2023gkd} introduce a novel approach where the student model first generates sequences, termed `self-generated sequences.' The student then learns by using feedback (i.e. output distribution) from teacher on these sequences. This method is particularly beneficial when the student model lacks the capacity to mimic teacher’s distribution.
Moreover, various LLM-quantization methods with distilling feature knowledge from teacher LLMs have been proposed~\cite{tao2022compression, liu2023llm, kim2023token}. These methods aim to preserve the original output distribution when quantizing the LLMs, ensuring minimal loss of performance.
Additionally, feature knowledge could serve as a potent source for multi-teacher knowledge distillation. \citet{timiryasov2023baby} leverages an ensemble of GPT-2 and LLaMA as teacher models to extract output distributions. Similarly, FuseLLM~\cite{wan2024knowledge} innovatively combines the capabilities of various LLMs through a weighted fusion of their output distributions, integrating them into a singular LLM. This approach has the potential to significantly enhance the student model's capabilities, surpassing those of any individual teacher LLM.

In summary, feature knowledge offers a more transparent alternative to black-box methods, allowing for deeper insight into and control over the distillation process. By utilizing feature knowledge from teacher LLMs, such as output distributions and intermediate layer features, white-box approaches enable richer knowledge transfer. 
While showing promise, especially in smaller models, its application is not suitable for black-box LLMs where internal parameters are inaccessible. Furthermore, student models distilled from white-box LLMs may underperform compared to their black-box counterparts, as the black-box teacher LLMs (e.g. GPT-4) tend to be more powerful.

\subsubsection{Feedback}

Most previous works predominantly focus on one-way knowledge transfer from the teacher to the student for imitation, without considering feedback from the teacher on the student's generation. The feedback from the teacher typically offers guidance on student-generated outputs by providing preferences, assessments, or corrective information. For example, a common form of feedback involves teacher ranking the student's generations and distilling this preference into the student model through Reinforcement Learning from AI Feedback (RLAIF)~\cite{bai2022constitutional}. Here is a generalized formulation for eliciting feedback knowledge:
\begin{equation}
\mathcal{D}^{(\text{fb})} = \{(x, y, \phi_{\text{fb}}(x, y;\theta_T)) | x \sim \mathcal{X}, y \sim p_S(y|x)\},
\end{equation}
where \(y\) denotes the output generated by the student model in response to \(x\), and \(\phi_\text{fb}(\cdot;\theta_T))\) represents providing feedback from teacher LLMs. This operation evaluates the student's output \(y\) given the input \(x\), by offering assessment, corrective information, or other forms of guidance. This feedback knowledge can not only be distilled into the student to also generate feedback (such as creating a student preference model) but, more importantly, enable the student to refine its responses based on the feedback. Various methods have been explored to elicit this advanced knowledge~\cite{bai2022constitutional, luo2023wizardmath, cui2023ultrafeedback, kwon2023reward, jiang2023lion, chen2023personalised, gu2023knowledge, agarwal2023gkd, chen2024improving, guo2023imitation, selfee2023, hong2023cyclealign, lee2023rlaif}.

Preference, as previously discussed, represents a notable form of feedback knowledge from teacher models.
Various knowledge of preferences could be distilled from teachers by prompting it with specific criteria. 
\citet{bai2022constitutional} introduce RLAIF for distilling harmlessness preferences from LLMs. This involves using an SFT-trained LLM to generate response pairs for each prompt, then ranking them for harmlessness to create a preference dataset. This dataset is distilled into a Preference Model (PM), which then guides the RL training of a more harmless LLM policy.
WizardMath~\cite{luo2023wizardmath} places emphasis on mathematical reasoning. They employ ChatGPT as teacher to directly provide process supervision and evaluate the correctness of each step in the generated solutions.
To scale up high-quality distilled preference data, \citet{cui2023ultrafeedback} develop a large-scale preference dataset for distilling better preference models, UltraFeedback. It compiles various instructions and models to produce comparative data. Then, GPT-4 is used to score candidates from various aspects of preference, including instruction-following, truthfulness, honesty and helpfulness. 

Beyond merely assessing student generations, teachers can also furnish extensive feedback on instances where students underperform. 
In Lion~\cite{jiang2023lion}, teacher model pinpoints instructions that pose challenges to the student model, generating new, more difficult instructions aimed at bolstering the student’s abilities. 
PERsD~\cite{chen2023personalised} showcases a method where teacher offers tailored refinement feedback on incorrect code snippets generated by students, guided by the specific execution errors encountered.
Similarly, SelFee~\cite{selfee2023} leverages ChatGPT to generate feedback and revise the student's answer based on the feedback.
In contrast, FIGA~\cite{guo2023imitation} revises the student's response by comparing it to the ground-truth response.
Furthermore, teacher model's distribution over the student's generations can itself act as a form of feedback. 
MiniLLM~\cite{gu2023knowledge} and GKD~\cite{agarwal2023gkd} present an innovative strategy wherein the student model initially generates sequences, followed by teacher model producing an output distribution as feedback. This method leverages the teacher's insight to directly inform and refine the student model's learning process.

\subsubsection{Self-Knowledge}
The knowledge could also be elicited from the student itself, which we refer to as \textit{Self-Knowledge}.
In this setting, the same model acts both as the teacher and the student, iteratively improving itself by distilling and refining its own previously generated outputs. This knowledge uniquely circumvents the need for an external, potentially proprietary, powerful teacher model, such as GPT-series LLMs. Furthermore, it allows the model to surpass the limitations or ``ceiling" inherent in traditional teacher-student methods. 
Eliciting self-knowledge could be formulated as: 
\begin{equation}
\mathcal{D}^{(\text{sk})} = \{ (x, y, \phi_\text{sk}(x, y)) | x \sim \mathcal{S}, y \sim p_S(y|I\oplus x) \},
\end{equation}
where \(\phi_\text{sk}(\cdot)\) is a generalized function that represents an additional process to the self-generated outputs \(y\), which could include but is not limited to filtering, rewarding, or any other mechanisms for enhancing or evaluating \(y\). It could be governed by external tools or the student itself $\theta_S$.
Recent research in this area has proposed various innovative methodologies to elicit self-knowledge, demonstrating its potential for creating more efficient and autonomous learning systems.~\cite{allen2020towards,wang2022self, sun2023principledriven, yang2023rlcd, jung2023impossible, huang-etal-2023-large, gulcehre2023reinforced, yuan2024selfrewarding, xu2023baize, zelikman2022star, chen2024selfplay, zheng2024kun, li2023selfalignment, zhao2024apt, singh2023beyond, chen2024grath, hosseini2024vstar}

A notable example of this methodology is Self-Instruct~\cite{wang2022self}, which utilizes GPT-3 for data augmentation through the \textit{Expansion} approach, generating additional data samples to enhance the dataset. This enriched dataset subsequently fine-tunes the original model. 
Other methods aim to elicit targeted knowledge from student models by modifying prompts, and leveraging these data for further refinement. 
In Self-Align~\cite{sun2023principledriven}, they find that models fine-tuned by Self-Instruct data tend to generate short or indirect responses. They prompt this model with verbose instruction to produce in-depth and detailed responses. Then, they employ context-distillation~\cite{askell2021general} to distill these responses paired with non-verbose instructions back to the model.
Similarly, RLCD~\cite{yang2023rlcd} introduces the use of contrasting prompts to generate preference pairs from an unaligned LLM, encompassing both superior and inferior examples. A preference model trained on these pairs then guides the enhancement of the unaligned model through reinforcement learning.
Several other approaches employ filtering methods to refine self-generated data. For example, Impossible Distillation~\cite{jung2023impossible} targets sentence summarization tasks, implementing filters based on entailment, length, and diversity to screen self-generated summaries. LMSI~\cite{huang-etal-2023-large} generates multiple CoT reasoning paths and answers for each question, and then retains only those paths that lead to the most consistent answer. 

Note that refined self-knowledge can be iteratively acquired as the student model continuously improves, further enhancing the student's capabilities. This is 
\citet{gulcehre2023reinforced} introduces a Reinforced Self-Training (ReST) framework that cyclically alternates between \texttt{Grow} and \texttt{Improve} stages to progressively obtain better self-knowledge and refine the student model. During the \texttt{Grow} stage, the student model generates multiple output predictions. Then, in the \texttt{Improve} stage, these self-generated outputs are ranked and filtered using a scoring function. Subsequently, the language model undergoes fine-tuning on this curated dataset, employing an offline RL objective.
Self-Play~\cite{chen2024selfplay} introduces a framework resembling iterative DPO, where the language model is fine-tuned to differentiate the self-generated responses from the human-annotated data. These self-generated responses could be seen as ``negative knowledge" to promote the student to better align with the target distribution.
Self-Rewarding~\cite{yuan2024selfrewarding} explores a novel and promising approach by utilizing the language model itself as a reward model. It employs LLM-as-a-Judge prompting to autonomously assign rewards for the self-generated responses. The entire process can then be iterated, improving instruction following and reward modeling capabilities.

\subsection{Distillation}
This section focuses on the 
methodologies for effectively transferring the elicited knowledge from teacher LLMs into student models.
We explore a range of distillation techniques, from the strategies that enhance imitation by \textit{Supervised Fine-Tuning}, \textit{Divergence} and \textit{Similarity}, to advanced methods like \textit{Reinforcement Learning} and \textit{Rank Optimization}, as shown in Figure~\ref{overall}.

\subsubsection{Supervised Fine-Tuning}
Supervised Fine-Tuning (SFT), or called Sequence-Level KD (SeqKD)~\cite{kim2016sequence}, is the simplest and one of the most effective methods for distilling powerful black-box LLMs. SFT finetunes student model by maximizing the likelihood of sequences generated by the teacher LLMs, aligning the student's predictions with those of the teacher. 
This process can be mathematically formulated as minimizing the objective function:
\begin{equation}
\mathcal{L}_{\text{SFT}} = \mathbb{E}_{x \sim \mathcal{X}, y \sim p_T(y|x)}\left[-\log p_S(y|x)\right],
\end{equation}
where \( y \) is the output sequence produced by the teacher model.
This simple yet highly effective technique forms the basis of numerous studies in the field. 
Numerous researchers have successfully employed SFT to train student models using sequences generated by teacher LLMs~\cite{alpaca, vicuna2023, wu2023laminilm, xu2023wizardlm, luo2023wizardmath}. Additionally, SFT has been explored in many self-distillation works
~\cite{wang2022self, huang2022large, xu2023baize, zelikman2022star}.
Due to the large number of KD works applying SFT, we only list representative ones here. More detailed works can be found in \S\ref{sec:skill}.

\subsubsection{Divergence and Similarity}

This section mainly concentrates on algorithms designed for distilling feature knowledge from white-box teacher LLMs, including distributions and hidden state features. These algorithms can be broadly categorized into two groups: those minimizing divergence in probability distributions and those aimed at enhancing the similarity of hidden states.

\begin{table}[t]
\centering
\scalebox{0.95}{
\begin{tabular}{cc}
\toprule
\textbf{Divergence Type} & \textbf{\( D(p, q) \) Function} \\
\midrule
Forward KLD & \( \sum p(t) \log \frac{p(t)}{q(t)} \) \\
\midrule
Reverse KLD & \( \sum q(t) \log \frac{q(t)}{p(t)} \) \\
\midrule
JS Divergence & \( \frac{1}{2} \left( \sum p(t) \log \frac{2p(t)}{p(t) + q(t)} + \sum q(t) \log \frac{2q(t)}{p(t) + q(t)} \right) \) \\
\bottomrule
\end{tabular}
}
\caption{Functional forms of \( D \) for various divergence types. $p$: reference }
\label{table:divergence-types}
\end{table}

\begin{table}[t]
\centering
\scalebox{0.94}{
\begin{tabular}{cc}
\toprule
\textbf{Similarity Function $\mathcal{L}_F$} & \textbf{Expression} \\
\midrule
L2-Norm Distance & \( \|\Phi_{T}(f_{T}(x, y)) - \Phi_{S}(f_{S}(x, y))\|_2 \) \\
\midrule
L1-Norm Distance & \( \|\Phi_{T}(f_{T}(x, y)) - \Phi_{S}(f_{S}(x, y))\|_1 \) \\
\midrule
Cross-Entropy Loss & \( -\sum \Phi_{T}(f_{T}(x, y)) \log(\Phi_{S}(f_{S}(x, y))) \) \\
\midrule
Maximum Mean Discrepancy & \(\text{MMD}(\Phi_{T}(f_{T}(x, y)), \Phi_{S}(f_{S}(x, y)))\) \\
\bottomrule
\end{tabular}}
\caption{Summary of similarity functions in knowledge distillation. }
\label{table:similarity-functions}
\end{table}

\begin{figure}[t]
    \centering
    \includegraphics[width=\linewidth]{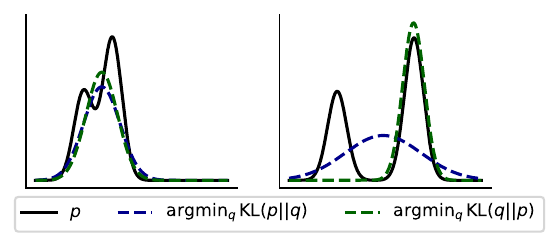}
    \caption{\textbf{Comparison of Forward and Reverse KL Divergences in Approximating a Target Distribution}. Forward KL divergence approach tends to cover all modes of the target distribution but is less precise, i.e. ``mode-covering" behavior. Reverse KL divergence method focuses predominantly on the most prominent mode, thereby exhibiting a ``mode-seeking" behavior.}
    \label{fig:kld}
\end{figure}

\vspace{2mm}
\pa{Divergence.}
Divergence-based methods minimize divergence between the probability distributions of the teacher and student models, represented by a general divergence function $D$:
\begin{equation}
L_{\text{Div}} = \underset{x \sim \mathcal{X}, y \sim \mathcal{Y}}{\mathbb{E}}\left[ D\left(p_T(y|x), p_S(y|x)\right) \right],
\end{equation}
The specific form of \( D \) varies depending on the type of divergence employed. Table \ref{table:divergence-types} outlines the functional forms of \( D \) for different divergence measures.
The commonly-used standard KD objectives essentially minimize the approximated forward Kullback-Leibler divergence (KLD) between the teacher and the student distribution~\cite{sanh2019distilbert, wen-etal-2023-f, timiryasov2023baby, liang2023less, chen2024knowledge}
, which forces $p_S$ to cover all the modes of $p_T$.
However, when a student model is unable to learn all modes of a highly complex teacher, the resultant ``mode-covering" behavior might cause the student to assign probability mass to tokens with low probability under the teacher's distribution (cf. Figure~\ref{fig:kld} blue curve). This mode-covering phenomenon can potentially lead to hallucinations and low-quality generations. 
Alternatively, mode-seeking divergences like reverse KL prioritize tokens where the teacher assigns high probabilities (cf. Figure~\ref{fig:kld} green curve). This approach can mitigate the risk of low-quality outputs, fostering more accurate generations. However, it often does so at the cost of reduced diversity.
\citet{gu2023knowledge} adopt reverse KL divergence to prevent students from overestimating low-probability regions of the teacher's distribution, employing Policy Gradient methods for optimization. 
Both \citet{agarwal2023gkd} and \citet{fd} assess the effect of different divergence functions in LLM distillation, finding the optimal divergence to be task-dependent. 
For instance, forward KL divergence is more suitable for tasks like Machine Translation, where the output has fewer modes or variations, while reverse KL divergence is preferable for tasks like dialogue generation and instruction tuning, which involve multiple modes and a wider range of potential responses.
Thus, the nature of the task significantly influences the selection of the divergence function for optimal performance.

\vspace{2mm}
\pa{Similarity.}
Similarity-based methods in knowledge distillation aim to align the hidden states or features of the student model with those of the teacher. These methods use various similarity metrics to measure and optimize the congruence of internal representations between the two models. The objective is to ensure that the student model not only produces similar outputs to the teacher but also processes information in a comparable manner. The formulation for a similarity-based objective might look like this:
\begin{equation}
\mathcal{L}_{\text{Sim}} = \underset{x \sim \mathcal{X}, y\sim \mathcal{Y}}{\mathbb{E}}\left[\mathcal{L}_{F}\left(\Phi_{T}\left(f_{T}(x,y)\right), \Phi_{S}\left(f_{S}(x,y)\right)\right)\right],
\end{equation}
where \( f_{T}(x,y) \) and \( f_{S}(x,y) \) are the feature maps of the teacher and student models, respectively. The transformation functions \( \Phi_{T} \) and \( \Phi_{S} \) are applied to these feature maps to ensure they are in the same shape, facilitating direct comparison. The similarity function \( \mathcal{L}_{F} \) is used to match these transformed feature maps. Table \ref{table:similarity-functions} shows common choices for \( \mathcal{L}_{F} \).
Few works have employed similarity-based methods in the KD of LLMs. Among them, \citet{liang2023less} propose Task-Aware Layer-Wise Distillation (TED), a method that utilizes task-aware filters. These filters are designed to selectively capture the most pertinent information for a specific task from the teacher model. The key objective is to minimize the discrepancy between the filtered representations in both teacher and student models. While similarity-based approaches are common in encoder-based LMs~\cite{sun2019patient, sun-etal-2020-mobilebert, jiao-etal-2020-tinybert,hou2020dynabert,zuo2022moebert,LiangHSZCCC21}, their application in LLM knowledge distillation is not as widespread. However, considering their effectiveness, we anticipate an increase in research exploring these methods for LLM distillation in the near future.

\subsubsection{Reinforcement Learning}

This section explores advanced methods of distilling knowledge into student models using reinforcement learning (RL). This approach is especially relevant for 
leveraging the feedback from teacher to train student models~
\cite{bai2022constitutional, cui2023ultrafeedback, luo2023wizardmath, agarwal2023gkd, chen2024improving, ma2023eureka, pang2023language, pmlr-v202-du23f}.
The RL-based distillation process typically involves two main stages:

\vspace{2mm}
\pa{Distilled Reward Model Training. }
The first stage involves training a reward model \( r_\phi \) using the feedback data \( \mathcal{D}^{(\text{fd})} \) generated by teacher LLMs. 
Preference data, as one of the typical feedback, is employed to train the student reward model~\cite{bai2022constitutional, cui2023ultrafeedback, lee2023rlaif, kim-etal-2023-aligning}. They usually consist of input-output pairs \((x, y_w, y_l)\). Here, \( y_w \) and \( y_l \) represent ``winning'' and ``losing'' outputs relative to the teacher's preferences. The loss function for the reward model is defined as:

\begin{equation}
\mathcal{L}_\text{RM}(r_\phi, \mathcal{D}^{\text{(fd)}}) = -\underset{(x, y_w, y_l) \sim \mathcal{D}^{\text{(fd)}}}{\mathbb{E}}\left[\log \sigma\left(r_\phi\left(x, y_w\right)-r_\phi\left(x, y_l\right)\right)\right]
\end{equation}

This formulation guides the reward model to correctly distinguish between more and less preferable outputs based on the teacher's criteria. 
Instead of learning the instance-level rewards, RLMEC~\cite{chen2024improving} adopts a different approach by training a generative reward model. It is trained on an erroneous solution rewriting data distilled from a teacher LLM. This distilled reward model can produce token-level rewards for RL training.

\vspace{2mm}
\pa{Reinforcement Learning Optimization.}
In the second stage, the student model, represented by a policy \( \pi_\theta \), is optimized to maximize the expected reward as per the trained reward model. Simultaneously, it minimizes the divergence from a reference policy \( \pi_{\mathrm{ref}} \), typically the initial policy of the student model trained by SFT, controlled by a factor \( \beta \). The RL objective is given by:

\begin{equation}
\label{eq:rl}
\max _{\pi_\theta} \underset{x \sim X, y \sim \pi_\theta(y \mid x)}{\mathbb{E}}\left[r_\phi(x, y)\right]-\beta {D}_{\mathrm{KL}}\left[\pi_\theta(y \mid x) \| \pi_{\mathrm{ref}}(y \mid x)\right]
\end{equation}
This RL framework not only ensures that the student model learns the explicit content from the teacher but also effectively adopts the teacher's preference patterns. The use of RL, particularly with the PPO~\cite{schulman2017proximal} algorithm, offers a robust mechanism for aligning the student model's outputs with the teacher.
Alternatively, the teacher LLM can also serve as the reward model to directly assign rewards during RL, circumventing the need for training a reward model~\cite{lee2023rlaif, kwon2023reward}. While this approach may exhibit superior performance, it comes at a higher computational cost compared to employing a smaller distilled reward model.

\subsubsection{Ranking Optimization}
Ranking optimization presents a stable and computationally efficient alternative to RL for injecting preference feedback into language models~\cite{rafailov2023direct, song2023preference, yuan2023rrhf}. This method, diverging from traditional RL approaches, directly incorporates ranking information into language models from a fixed preference dataset during fine-tuning. Intuitively, it directly updates policy to increase the relative likelihood of preferred over less favored responses. 
This direct optimization of preferences, without the need for sampling outputs, makes the process more stable and efficient. 
Recently, some works have been proposed to explore using ranking optimization to distill teacher's preferences into student models~\cite{tunstall2023zephyr, hong2023cyclealign, yuan2024selfrewarding}.

Zephyr~\cite{tunstall2023zephyr} utilizes Direct Preference Optimization (DPO)~\cite{rafailov2023direct} to distill the preference alignment in teacher LLMs. DPO streamlines the objective of reinforcement learning (as in Eq.~\ref{eq:rl}), which involves reward maximization with a KL-divergence constraint, into a single-stage policy training. Specifically, DPO's training goal is to maximize the following expectation:
\begin{equation}
 \underset{(x,y_{w},y_{l})\sim D^{\text{(fd)}}}{\mathbb{E}}\left[\log\sigma\left(\beta\log\frac{\pi_\theta(y_{w}|x)}{\pi_{\mathrm{ref}}(y_{w}|x)}-\beta\log\frac{\pi_\theta(y_{l}|x)}{\pi_{\mathrm{ref}}(y_{l}|x)}\right)\right],
\end{equation}
where \(y_w\) is preferred over \(y_l\) according to the teacher LLM.
Hong et al. (2023)~\cite{hong2023cyclealign} adopt two ranking-based optimization objectives, Rank Responses to align Human Feedback (RRHF)~\cite{yuan2023rrhf} and Preference Ranking Optimization (PRO)~\cite{song2023preference}, for preference distillation. 
RRHF~\cite{yuan2023rrhf} focuses on a ranking loss defined as:
\begin{equation}
    \mathcal{L}_\text{RRHF}=\sum_{r_{i} < r_{j}}\operatorname*{max}(0,p_{i}-p_{j}),
\end{equation}
where \(r_i\) and \(r_j\) are the reward scores assigned by the teacher LLM for responses \(y_i\) and \(y_j\), respectively, and \(p_i\), \(p_j\) are their corresponding conditional log probabilities under the policy \(\pi_\theta\). This approach emphasizes direct comparison and ranking of responses based on the teacher's preferences.
PRO~\cite{song2023preference} expands the concept of pairwise comparison to handle preference rankings of any length. For a given instruction $x$ and a sequence of responses ordered by teacher preference as $y_1 \succ y_2  \succ  ... \succ y_n$, the RPO training objective is:
\begin{equation}
    {\mathcal{L}}_{\mathrm{PRO}}=-\sum_{k=1}^{n-1}\log{\frac{\exp\left(p_k\right)}{{\sum^{n}_{i=k}}\exp\left(p_i\right)}},
\end{equation}
where $p_k$ represents the conditional log probabilities for $y_k$ under the student policy \(\pi_\theta\).
By iteratively contrasting the likelihood of generating responses, PRO optimizes the student LM to prioritize the most preferred response while progressively ranking the rest in the order of diminishing preference.

% \textbf{Papers}: Ensemble~\cite{timiryasov2023baby}, Context-distillation~\cite{padmanabhan2023propagating} 

\begin{table*}[!h]
    \centering
    \resizebox{0.99\textwidth}{!}{
    \begin{tabular}{lcccccc}
    \toprule
    \textbf{Methods} & \textbf{Skill} &  \textbf{Seed Knowledge} &  \textbf{Teacher LLM}&\textbf{Student Model}&\textbf{Knowledge Elicitation} & \textbf{Objective} \\
    \midrule 
    \multicolumn{7}{c}{ \textbf{\emph{Context Following}}}\\
    \midrule
    Self-Instruct~\cite{wang2022self} & IF  & 175 human-curated tasks & GPT3 & GPT3 & Expansion + Self-Knowledge & SFT \\
    Alpaca~\cite{alpaca} & IF  & 175 human-curated tasks & GPT3 & LLaMA & Expansion + Self-Knowledge & SFT \\
    LaMini-LM~\cite{wu2023laminilm} & IF  & \begin{tabular}[c]{@{}l@{}} 3.5K Wikipedia Categories + \\ \quad\quad\quad Mixed Dataset \end{tabular} & ChatGPT & Various Models & Expansion & SFT \\
    WizardLM~\cite{xu2023wizardlm} & IF  & Alpaca Data & ChatGPT & LLaMA & Expansion & SFT \\
    Lion~\cite{jiang2023lion}& IF  & Alpaca Cata & ChatGPT & LLaMA & Labeling + Expansion + Feedback & - \\
    BabyLlama~\cite{timiryasov2023baby}& IF  & 10M-word BabyLM dataset & GPT-2 + small LLaMA & 58M-parameter LLaMA & Feature & D\&S \\
    MiniLLM~\cite{gu2023knowledge}& IF  & Dolly Dataset & GPT2 + OPT + LLaMA & GPT2 + OPT + LLaMA & Feature & D\&S \\
    Self-Align~\cite{sun2023principledriven}& IF  & Human-written Principles & LLaMA & LLaMA & Expansion + Self-Knowledge & SFT \\
    Self-Rewarding~\cite{yuan2024selfrewarding}& IF  & Human-written Samples & LLaMA & LLaMA & Self-Knowledge & SFT + RL \\
    STaR~\cite{zelikman2022star}& IF  &  Arithmetic + CommonsenseQA + GSM8K & GPT-J & GPT-J & Self-Knowledge & SFT \\
    Llama-GPT4~\cite{peng2023instruction}& IF  & Alpaca Dataset & GPT4 & LLaMA & Labeling & SFT \\
    Reflection-Tuning~\cite{li2023reflectiontuning} & IF  & Alpaca/WizardLM Dataset & ChatGPT & LLaMA & Labeling & SFT \\
    Selective Reflection-Tuning~\cite{Li2024SelectiveRS} & IF  & Alpaca/WizardLM Dataset & ChatGPT & LLaMA & Labeling & SFT \\
    Vicuna~\cite{vicuna2023} & IF/MD  & Human Conversation & ChatGPT + GPT4 & LLaMA & Labeling & SFT \\
    Koala~\cite{koala_blogpost_2023} & IF/MD  & Human Conversation & ChatGPT & LLaMA & Labeling & SFT \\
    Baize~\cite{xu2023baize} & IF/MD  & Quora + Stack Overflow & ChatGPT & LLaMA & Expansion + Self-Knowledge & SFT \\ 
    UltraChat~\cite{ding2023enhancing} & IF/MD  & Wikidata + Text Material + C4 & ChatGPT & LLaMA & Curation & SFT \\ 
    Orca~\cite{mukherjee2023orca} & IF/TP  & FLAN-v2 & ChatGPT + GPT4 & LLaMA & Labeling & SFT \\
    Orca2~\cite{mitra2023orca2} & IF/TP  & FLAN-v2 + Few-Shot/Math/Synthetic & GPT4 & LLaMA & Labeling & SFT \\
    SelFee~\cite{selfee2023} & IF/TP  & Human Conv, Flan/Code/Math Collection & ChatGPT & LLaMA & Labeling & SFT \\
    CoT-Distill~\cite{hsieh2023distilling}& IF/TP  & e-SNLI + ANLI + CQA + SVAMP & PaLM & T5 & Labeling & SFT \\
    KnowPAT~\cite{zhang2023knowledgeable} & IF/TP  & CPKG + QA Data & ChatGPT + ChatGLM + Vicuna-7B & LLaMA & Labeling & SFT \\
    DEBATunE~\cite{li2024llms} & IF/TP  & Controversial Topics & ChatGPT & LLaMA & Labeling & SFT \\
    Phi-1~\cite{gunasekar2023textbooks}  & IF/Code  &  - & GPT3.5 & phi-1 & Curation & SFT \\
    Phi-1.5~\cite{li2023textbooks1.5} & IF/Code  &  20k Topics from Web & GPT3.5 & phi-1 & Curation + Labeling & SFT \\
    SAIL~\cite{luo2023sail} & IF/RAG  & Alpaca Data + Web Content & GPT4 & LLaMA & Label & SFT \\ 
    KARD~\cite{kang2023knowledgeaugmented} & IF/RAG  & MedQAUSMLE & ChatGPT & T5 + OPT & Label & SFT + D\&S \\ 
    Self-RAG~\cite{asai2023self} & IF/RAG  & Open-Instruct & GPT4 & LLaMA & Labeling & SFT \\ 

    \midrule
    \multicolumn{7}{c}{ \textbf{\emph{Alignment}}}\\
    \midrule
    OpenChat~\cite{wang_openchat_2023} & IF/Preference  & Human Conversation & ChatGPT + GPT4 & LLaMA & Labeling & SFT + RL \\
    Zephyr~\cite{tunstall2023zephyr} & IF/Preference  & Mixed Datasets & GPT4 & Mistral & Labeling + Feedback & SFT + RO \\
    ALMoST~\cite{kim-etal-2023-aligning} & IF/Preference  & Human-written Prompts & LLaMA & LLaMA & Expansion + Labeling & SFT + RL \\
    RLCD~\cite{yang2023rlcd} & IF/Preference  & Human-written Prompts & LLaMA & LLaMA &  Labeling & SFT + RL \\
    RLAIF~\cite{lee2023rlaif} & IF/Preference  & Human-written Prompts & PaLM 2 & PaLM 2 & Labeling + Feedback & RL \\
    GPT3 Reward~\cite{kwon2023reward} & Preference  &  Human-written Prompts & GPT3 & GPT3 & Labeling & RL \\
    ILF~\cite{scheurer2023training} & Preference  &  Task-specific Datasets & GPT3 + FeedME & GPT3 & Labeling & RL \\
    ULTRAFEEDBACK~\cite{cui2023ultrafeedback} & Preference  & Mixed Datasets & GPT4 & LLaMA & Labeling & RL \\
    Constitutional AI~\cite{bai2022constitutional} & Preference/Value  &  Human-written Prompts & Self-defined Student Model & Self-defined Model & Labeling + Expansion + Feedback & SFT + RL \\
    SANDBOX~\cite{liu2023training} & Value  & Simulation & \begin{tabular}[c]{@{}l@{}} text-davinci-002/-003 +\\ \quad\quad GPT4 + ChatGPT \end{tabular} & LLaMA & Data Curation & SFT + RL \\
    
    \midrule
    \multicolumn{7}{c}{ \textbf{\emph{Agent}}}\\
    \midrule
    Toolformer~\cite{schick2023toolformer} & Tool  & CCNet & GPT-J & GPT-J & Labeling & SFT \\
    Graph-ToolFormer~\cite{zhang2023graph} & Tool  & Mixed Graph Dataset & ChatGPT & GPT-J + LLaMA & Labeling & SFT \\
    Gorilla~\cite{patil2023gorilla} & Tool  & Online API Documentation & GPT4 & LLaMA & Expansion & SFT \\
    GPT4Tools~\cite{yang2023gpt4tools} & Tool  & Image Content & ChatGPT & LLaMA & Curation + Expansion & SFT \\
    ToolAlpaca~\cite{tang2023toolalpaca} & Tool  & Public-apis Repository & ChatGPT & LLaMA & Curation & SFT \\
    ToolLLM~\cite{qin2023toolllm} & Tool  & Real-world APIs & ChatGPT & LLaMA & Curation & SFT \\
    MLLM-Tool~\cite{wang2024mllmtool} & Tool  & HuggingFace Model Cards & GPT4 & LLaMA & Curation & SFT \\
    FireAct~\cite{chen2023fireact} & Planning  & Mixed QA Dataset & GPT4 & LLaMA & Labeling & SFT \\
    AgentTuning~\cite{zeng2023agenttuning} & Planning  & 6 Agent Tasks & GPT4 + ChatGPT& LLaMA & Labeling + Expansion & SFT \\
    Lumos~\cite{yin2023lumos} & Planning  & Mixed Interactive Tasks & GPT4 & LLaMA & Labeling & SFT \\
    AUTOACT~\cite{qiao2024autoact} & Planning  & Mixed QA Tasks & LLaMA & LLaMA & Labeling & SFT \\

    \midrule
    \multicolumn{7}{c}{ \textbf{\emph{NLP Task Specialization}}}\\
    \midrule
    AugGPT~\cite{dai2023auggpt} & NLU  & Amazon/Symptoms/PubMed20k Dataset & ChatGPT & BERT & Label & SFT \\
    TDG~\cite{he-etal-2023-targeted} & NLU  & SST + QQP + MNLI & GPT3 & BERT & Expansion & SFT \\
    SunGen~\cite{DBLP:conf/iclr/GaoPLXY0ZLLK23} & NLU  & Text Classification Tasks & GPT2 & DistilBERT & Curation & SFT \\
    UDG~\cite{wang2021zerolabel} & NLU  &  NLU Tasks & GPT3 & BERT & Expansion & SFT \\
    InheritSumm~\cite{xu-etal-2023-inheritsumm} & NLG  & Pile + ArXiv + CNN/DM + WikiHow & GPT3.5 & ZCode++  & Label & SFT \\
    DIMSUM+~\cite{jung2023impossible} & NLG  & None & GPT2 + CTRL + BioGPT & T5 & Curation + Self-Knowledge & SFT \\
    Genie~\cite{yehudai2024genie} & NLG  & ELI5 + ASQA + NQ + CNN/DM & Falcon + LLaMA & FLAN + LLaMA & Label & SFT \\
    GKD~\cite{agarwal2023gkd} & NLG/NLU/IF  & XSum+WMT14 en-de+GSM8K+FLAN2021 & T5-XL & T5 & Feature + Feedback & D\&S + RL \\
    QUILL~\cite{srinivasan-etal-2022-quill} & IR  & IR Datasets & T5 & 4-layer Transformer & Internal Knowledge & D\&S \\
    RankVicuna~\cite{pradeep2023rankvicuna} & IR  & IR Datasets & ChatGPT & LLaMA & Labeling & SFT \\
    RankZephyr~\cite{pradeep2023rankzephyr} & IR  & IR Datasets & ChatGPT + GPT4 & Mistral & Labeling & SFT \\
    NDR~\cite{10.1145/3604915.3608829} & Recommendation  & Recommendation Datasets & GPT3 & MPnet-110M & Labeling & SFT \\
    InstrcutRec~\cite{zhang2023recommendation} & Recommendation  & 39 instruction templates & ChatGPT & Flan-T5 & Expansion + Self-Knowledge & SFT \\
    ONCE~\cite{liu2023once} & Recommendation  & Recommendation Dataset & ChatGPT & LLaMA & Labeling & SFT \\
    PandaLM~\cite{wang2023pandalm} & Evaluation  & Alpaca Data & ChatGPT & LLaMA & Labeling & SFT \\
    Prometheus~\cite{kim2023prometheus} & Evaluation  & 50 Seed Rubrics & GPT4 & LLaMA & Labeling & SFT \\
    InstructScore~\cite{xu-etal-2023-instructscore} & Evaluation  & Mixed Dataset & GPT4 & LLaMA & Labeling & SFT \\
    WizardMath~\cite{luo2023wizardmath} & Math  & GSM8k + MATH & ChatGPT & LLaMA & Expansion + Feedback & SFT + RL \\
    Mammoth~\cite{yue2023mammoth}& Math/TP  & Mixed Math Dataset & GPT4 & LLaMA & Labeling & SFT \\
    Mixed Distill~\cite{chenglin2023mixed}& Math/TP  & SVAMP + GSM8K + ASDIV + StrategyQA & ChatGPT & LLaMa  & Labeling & SFT \\
    WizardCoder~\cite{luo2023wizardcoder} & Code  & Code Alpaca Data & ChatGPT & StarCoder & Expansion & SFT \\
    Magicoder~\cite{wei2023magicoder}& Code  & Existing Source Codes & ChatGPT & LLaMa  & Curation & SFT \\
    WaveCoder~\cite{yu2024wavecoder}& Code  & Existing Source Codes & GPT4 & LLaMa  & Curation & SFT \\
    Code Alpaca~\cite{codealpaca} & Code  & Code Instructions & ChatGPT & LLaMA & Expansion + Self-Knowledge & SFT \\
    Code Llama~\cite{rozière2023code} & Code  & Human-written Instructions & LLaMA & LLaMA & Expansion + Self-Knowledge & SFT \\
    Code Clean~\cite{DBLP:journals/corr/abs-2311-14904} & Code  & Code Datasets & ChatGPT & LLaMA & Labeling & SFT \\
    \midrule
    \multicolumn{7}{c}{ \textbf{\emph{Multi-Modality}}}\\
    \midrule
    LLaVA~\cite{liu2023visual} & Vision-Language  & COCO & GPT4 & LLaMA & Labeling & SFT \\
    SVIT~\cite{zhao2023svit} & Vision-Language  & Visual Genome + COCO & GPT4 & LLaMA & Labeling & SFT \\
    LVIS-Instruct4V~\cite{wang2023believe} & Vision-Language  & LVIS & GPT4V & LLaMA & Labeling & SFT \\
    LLaVAR~\cite{zhang2023llavar} & Vision-Language  & LAION & GPT4 & LLaMA & Labeling & SFT \\
    Macaw-LLM~\cite{lyu2023macaw} & Multiple Modalities  & Image/Video with Caption  & ChatGPT & LLaMA & Labeling & SFT \\
    MIMIC-IT~\cite{li2023mimicit} & Multiple Modalities  & Image/Video Dataset & ChatGPT & LLaMA & Labeling & SFT \\
    ChatBridge~\cite{zhao2023chatbridge} & Multiple Modalities  & Task-Specific/Multimodal-Chat Data & GPT4 + ChatGPT & LLaMA & Labeling & SFT \\
    \bottomrule
    \end{tabular}}
    \caption{A summary of skill distillation works. IF: Instruction Following, MD: Multi-turn Dialoue, TP: Think Pattern, RAG: Retrieval-Augmented Generation, NLU: Natural Language Understanding, NLG: Natural Language Generation, IR: Information Retrieval, SFT: Supervised Fine-Tuning, D\&S: Divergence and Similarity, RL: Reinforcement Learning, RO: Ranking Optimization.}
    \label{tab:my_label}
\end{table*}

\section{Skill Distillation}\label{sec:skill}

Building upon the foundation laid out in Section~\ref{sec:kd} about eliciting knowledge and distillation algorithms, we shift our focus to how these techniques facilitate the distillation of specific skills in LLMs. 
Our exploration will encompass a diverse range of skills exhibited by LLMs, including \textit{Context Following}, \textit{Alignment}, \textit{Agent}, \textit{NLP Task Specialization} and \textit{Multi-Modality}.
\textit{Context Following} focuses on the student's ability to comprehend and respond effectively to input information. \textit{Alignment} delves into the student's capability to align its output with the teacher's responses. Moving forward, \textit{Agent} underscores the autonomous nature of language models. \textit{NLP Task Specialization} highlights the LLM's versatility in specializing across various Natural Language Processing tasks, demonstrating its adaptability. Finally, \textit{Multi-Modality} encompasses the knowledge transfer from teacher LLMs to multi-modal models. Table~\ref{tab:my_label} summarizes the representative works, encompassing details such as the skills involved, seed knowledge, teacher LLM, student model, knowledge elicitation method, and training objectives.

\subsection{Context Following}

This part concentrates on the distillation of context following skills from LLMs. This process involves transferring the ability of LLMs to handle a variety of complex contexts — such as few-shot demonstrations, intricate instructions, dialogue history, and retrieval-augmented information — into smaller models. Many research efforts in this domain aim to imbue smaller models with these sophisticated, context-following capabilities. Our discussion here will dissect this facet of skill distillation, categorizing it based on different types of context and elaborating on how each is distilled and incorporated into smaller, efficient models.

\subsubsection{Instruction Following}

Instruction-following capacity enables LLMs to understand and follow user-given instructions. This ability significantly enhances human-AI interaction, allowing for seamless understanding and execution of tasks as directed by users.
A primary method for acquiring this skill involves constructing instruction-like prompt-response pairs and employing Supervised Fine Tuning (SFT) for model training. 
Data for this purpose can be manually curated by human experts or transformed from existing NLP tasks into instructional formats with templates, such as prefacing machine translation data with \textit{"Translate this sentence to Spanish:"}.
However, these approaches have limitations. Manual data creation is labor-intensive, while template-based transformation lacks diversity in instructions and may not align well with natural human input.
LLMs like GPT-4 offer an efficient alternative for creating diverse and controlled SFT data by their capabilities of in-context learning and instruction following. 
Most relevant works use OpenAI's GPT series models to generate prompt-response data pairs and then train the student LLMs by supervised fine-tuning~\cite{wang2022self, alpaca, vicuna2023, wu2023laminilm, xu2023wizardlm, mukherjee2023orca, mitra2023orca2, luo2023wizardmath, peng2023instruction}.

\vspace{2mm}
\pa{Basic Instructions.}
Self-Instruct~\cite{wang2022self} leverages the in-context learning capability of GPT-3 to expand a seed pool of 175 tasks to 52K task-agnostic instructions, ensuring a broad spectrum of general instructions. Additionally, a filtering and post-processing stage is introduced to eliminate redundant or similar instructions. Notably, through training with this enriched dataset, GPT-3 acquires the ability to follow instructions, enabling it to perform comparably to InstructGPT in zero-shot instruction tasks and when provided with expert-written instructions for novel tasks.
Based on the self-instruct method, \citet{alpaca} train an Alpaca model using the Llama 7B model on 52K instruction-following demonstrations, generated in a similar style as self-instruct but utilizing the more robust text-davinci-003 model. 
To enhance the diversity of instructional data, \citet{wu2023laminilm} introduce a technique known as \textit{Topic-Guided Instruction Generation}. This method involves gathering 3.5K common topics from Wikipedia to serve as guidance during the generation process. 

\vspace{2mm}
\pa{Complex Instructions.}
Some works promote students to solve more complex instructions~\cite{xu2023wizardlm, luo2023wizardmath, luo2023wizardcoder, guo2023instruction}.
According to \citet{xu2023wizardlm}, instruction datasets derived from human-written seeds often exhibit low to moderate complexity. To enhance the complex instruction-following capabilities of smaller models, WizardLM~\cite{xu2023wizardlm} introduces \textit{Evol-Instruct}. This method gradually transforms instructions into more complex forms through a multi-step evolution process, focusing on both increasing difficulty levels and expanding the diversity of topics. They conducted four rounds of evolution using the OpenAI ChatGPT API, resulting in a dataset of 250k complex instructions. Subsequently, they trained the LLaMA 7B model, referred to as WizardLM, on this dataset. In the high-difficulty section of test instructions, WizardLM even outperformed ChatGPT, achieving a win rate 7.9\% higher than ChatGPT. \citet{zhao2023preliminary} further conduct preliminary studies revealing the effectiveness of increasing instruction complexity. 
Instruction Fusion~\cite{guo2023instruction} further uses teacher LLMs to increase the complexity by fusing two distinct evolved instructions.
Furthermore, this concept of ``evolving" instructions has been extended to distill specific skills such as coding~\cite{luo2023wizardcoder} and mathematics~\cite{luo2023wizardmath}.

\vspace{2mm}
\pa{Human Instructions.}
In contrast to works that rely on generating instructions from ChatGPT, which may lack diversity and have gaps with real human instructions, Vicuna~\cite{vicuna2023} and Koala~\cite{koala_blogpost_2023} showcase impressive performance by using human conversations and natural instructions from community-contributed conversations. These conversations, found in platforms like ShareGPT, provide a forum for users to share their interactions with ChatGPT. It's important to note, however, that models trained on such natural conversations might mimic the style but may not fully capture the reasoning process of the original teacher~\cite{gudibande2023false, mukherjee2023orca}.

\vspace{2mm}
\pa{System Instructions.}
To encourage student models to learn the reasoning process, Orca and Orca 2~\cite{mukherjee2023orca, mitra2023orca2} enhance the \textit{{prompt, response}} data pairs by introducing a \textit{system message} (e.g., "explain like I’m five, think step-by-step") to encourage student models to grasp the reasoning process. This \textit{system message} prompts GPT-4 to provide explanation traces that elucidate the teacher's reasoning process.
Orca 2~\cite{mitra2023orca2} further trains the student model to identify the most effective solution strategy for each task, guided by Orca's performance. This approach significantly improves the ability of smaller models to follow instructions that involve reasoning.

\vspace{2mm}
\pa{High-Quality Instructions.}
As demonstrated in \citet{zhou2023lima} and \cite{Li2024SuperfilteringWD}, the data quality is crucial for instruction following training. 
UltraChat~\cite{ding2023enhancing} distills large-scale data with high-quality and diverse instructions from teacher LLMs by various meta-information. The UltraLLaMA model, fine-tuned on this data, consistently surpasses other open-source models.
The Phi series models~\cite{gunasekar2023textbooks, li2023textbooks1.5, Marah2023phi2} prioritize data quality and employ synthetic methods to generate data of ``textbook quality" to enhance the learning experience for smaller models. Notably, Phi exhibits the ability to follow instructions effectively even without specific instruction fine-tuning. What's particularly remarkable is that Phi-2, with just 2.7 billion parameters, outperforms Mistral and Llama-2 models with 7B and 13B parameters across various benchmark evaluations.

\vspace{2mm}
\pa{Improved Instructions.}
Another line of work focuses on improving the quality of existing instruction data, including both the improvement of instruction and corresponding response. SelFee~\cite{selfee2023} utilizes the ChatGPT to iteratively improve the quality of responses. ExpertLLaMA~\cite{xu2023expertprompting} improves the quality of responses by augmenting vanilla instructions with specialized Expert Identity descriptions. Reflection-Tuning~\cite{li2023reflectiontuning} improves both the instruction and response sequentially by reflecting on specific criteria. DEITA~\cite{liu2023makes} proposes to enhance and score instructions in three directions including complexity, quality, and diversity to get high-quality distillation data. MUFFIN~\cite{lou2023muffin} proposes to scale the instruction according to the input by diversifying these tasks with various input facets. Selective Reflection-Tuning~\cite{Li2024SelectiveRS} first involves the student model in the data improvement pipeline with a novel student-selection module, in which the student model is able to decide the data learn from.

In summary, distilling instruction data from teachers presents a promising avenue for training cheap and reproducible instruction-following language models. Current small models have made strides in enhancing various aspects of instruction-following ability, like diversity, complexity and explanation. 
However, student models trained on instruction data expanded by ChatGPT often mimic ChatGPT's style without replicating its factual accuracy~\cite{gudibande2023false}. Achieving a more capable instruction-following capability requires a stronger teacher LLM~\cite{gudibande2023false} and access to diverse, high-quality instruction data, such as the one used in Orca~\cite{mukherjee2023orca, mitra2023orca2}, which incorporates extensive task instructions from the Flan 2022 Collection~\cite{longpre2023flan}.

\subsubsection{Multi-turn Dialogue}

While instruction following focuses on single-instance command execution, multi-turn dialogue extends this to comprehend and maintain context through ongoing interactions. This skill is vital for models to engage meaningfully in human-like conversations and respond coherently over successive dialogue turns.
Some works have been dedicated to train to small chat models by distilling multi-turn knowledge from teacher LLMs~\cite{vicuna2023, xu2023baize, ding2023enhancing, li2023camel, wang_openchat_2023, tunstall2023zephyr}.

ShareGPT serves as a platform for users to share their conversations with ChatGPT, offering a vast repository of multi-turn conversations readily available. Some small chat models are trained using this data to acquire the capability for engaging in multi-turn dialogues~\cite{vicuna2023, selfee2023, wang_openchat_2023}.
For example, Vicuna~\cite{vicuna2023} is a chat model exclusively trained on ShareGPT data. Despite its sole training source being ShareGPT, Vicuna achieves a high MT-Bench~\cite{DBLP:journals/corr/abs-2306-05685} score assigned by GPT-4\footnote{MT-Bench: a multi-turn question set, where the generations of models are evaluated by LLM, like GPT-4.}.
In the study conducted by \citet{wang_openchat_2023}, GPT-3.5 and GPT-4 are employed to generate mixed responses using ShareGPT data. They assign higher rewards to responses generated by GPT-4, aiming to incentivize student models to produce high-quality responses.
Additionally, \citet{selfee2023} enhance the quality of multi-turn data from ShareGPT by generating self-feedback on model responses and iteratively refining the responses based on the received feedback.

To enhance the multi-turn capabilities of student models, another line of research focuses on expanding conversational datasets through self-chat and using them to train smaller models~\cite{xu2023baize, ding2023enhancing, tunstall2023zephyr}.
For instance, \citet{xu2023baize} initiate their work by using questions sourced from Quora and Stack Overflow as seeds, resulting in the collection of 111.5k dialogues through self-chat. Subsequently, they employ parameter-efficient tuning to train a chat model named Baize.
\citet{ding2023enhancing} first construct a significantly larger dataset called UltraChat, comprising 1.5 million high-quality multi-turn dialogues. They achieve this by distilling instructions and dialogues from ChatGPT. Notably, UltraChat encompasses a wide range of topics and instructions. Building upon the UltraChat dataset, they fine-tune a LLaMA model, resulting in the creation of a powerful chat model known as UltraLLaMA. UltraLLaMA consistently outperforms other open-source chat models, including Vicuna and Baize.
Furthermore, UltraChat is employed in conjunction with an AI preference-aligned chat model named Zephyr~\cite{tunstall2023zephyr}. Zephyr enhances intent alignment through the application of distilled direct preference optimization (dDPO).

\subsubsection{RAG Capbility}

LLMs are known to lack the ability to utilize up-to-date knowledge, and often produce responses containing factual inaccuracies due to their sole reliance on the parametric knowledge.
Retrieval-Augmented Generation (RAG) is a promising technique to decrease this issue. 
Handling the augmented context of retrieved information is also a non-trivial skill of LLMs. 
Several approaches to distill RAG capabilities have been proposed~\cite{kang2023knowledge, luo2023sail, asai2023self}.

SAIL~\cite{luo2023sail} starts by retrieving search results for each training case using search APIs, creating search-augmented instructions that include both the instruction and grounding information. To encourage the language model to prioritize informative retrieval results, they input each retrieved passage along with the ground truth response into the entailment model to label each retrieval result for relevance. Subsequently, the search-augmented instructions and relevance labels are fed into teacher LLMs (like GPT-4) for generating responses. Following fine-tuning on this training set, the student model becomes proficient at denoising search results and generating accurate responses.
KARD~\cite{kang2023knowledgeaugmented} distills rationales $r$ from the teacher LLM in response to questions $x$. These rationales are then utilized to train two models: a student LM and a Reranker.
For training the student LM, the rationales serve as a means to retrieve relevant knowledge $d$, and the student LM is subsequently fine-tuned using the rationales alongside questions and knowledge. However, during inference, only questions are available. To address this, the Reranker is trained to mimic how the retriever scores passages with the rationale by minimizing the KL divergence between $Retriever(d|r)$ and $Reranker(d|x)$.
However, the integration of a fixed number of passages in language models, without considering their necessity or relevance, can reduce versatility and lead to the generation of unhelpful responses.
To equip student LMs with adaptive RAG capabilities, Self-Rag~\cite{asai2023self} distills this adaptive ability from teacher LLMs into a small critic model. This critic model determines whether retrieval is necessary and evaluates the quality of the retrieved results by generating `reflection tokens.' For instance, Self-Rag initiates the retrieval operation when generating the reflection token \boxed{Retrieve}. To distill this critic data, GPT-4 is prompted to assess the need for retrieval using few-shot demonstrations $I$, the task input $x$, and output $y$ to predict a reflection token $r$ as follows: $p(r|I,x,y)$.

\subsection{Alignment}

\subsubsection{Thinking Pattern}
Most existing methods mainly focus on directly aligning the direct responses of the student models to the responses of teacher models~\cite{alpaca}. Though effective, these models might suffer the problems that they tend to learn to imitate the response style of the teacher models, but not the reasoning process~\cite{mukherjee2023orca}. Thus in order to better distill from the teacher models, methods are proposed that not only imitate the pure responses but some novel thinking patterns~\cite{selfee2023, mukherjee2023orca, mitra2023orca2, wang2023making, cheng2023adapting, zhang2023knowledgeable}.

Motivated by the effectiveness of LLMs in generating their own feedback without relying on external models~\cite{schick2022peer, madaan2023selfrefine, saunders2022selfcritiquing}, SelFee~\cite{selfee2023} proposes to train a model that has been fine-tuned to continuously revise its own answer until it provides a high-quality response in a single inference. During training, it utilizes both the final response and feedback chain as the fitting target. This pattern, response with the revision process, shows a promising performance gain. Following SelFee, Reflection-Tuning~\cite{li2023reflectiontuning, Li2024SelectiveRS} also utilizes the reflection process as the learning pattern. 
Noticing the lack of reasoning imitation of the previous methods, Orca~\cite{mukherjee2023orca} first proposes Explanation tuning, which aims to learn the reasoning steps, including explanation traces, step-by-step thought processes, and other complex instructions, from the teacher model, rather than just the vanilla styles. Extensive experiments verify the effectiveness of distilling with this thinking pattern. The following Orca2~\cite{mitra2023orca2} further presents to equip the student models with the ability to utilize different solution strategies for different tasks, motivated by the capability discrepancies between the smaller and larger models. By employing this training pattern, the student models are able to gain a better reasoning ability. 
Besides learning with the corresponding revision or reflection process, another thinking pattern that recently appeared is generating both responses and preferences. 
\citet{zhang2023knowledgeable} propose to learn both the knowledge and corresponding preference for domain-specific QA with LLMs. 
Recently, DEBATunE \cite{li2024llms} proposes to improve the controllability of LLMs in generating statements on controversial topics. 
By engaging two agents in a structured multi-round debate on controversial topics, salient and in-depth statements can be obtained and further distilled into the student models.

\subsubsection{Preference}
The previously mentioned methods primarily focus on the basic capability of student models to produce outcomes that are strictly accurate but may not align with human preferences, reaching alignment at this level enables these models to aid in various tasks without meeting higher-level demands. Early methods mainly utilize human feedback for the alignment of human preferences~\cite{ziegler2019fine, stiennon2020learning, wu2021recursively, ouyang2022training, bai2022training, köpf2023openassistant, yuan2023rrhf}. However, obtaining human feedback is costly and labor-intensive, thus methods that learn from AI feedback are also proposed to align with human preferences~\cite{bai2022constitutional, kwon2023reward, scheurer2023training, kim-etal-2023-aligning, roit-etal-2023-factually, yang2023rlcd, lee2023rlaif, tunstall2023zephyr, cui2023ultrafeedback, wang2023openchat}. 

The concept of RLAIF, introduced by \citet{bai2022constitutional}, involves the integration of preferences labeled by LLMs with those labeled by humans. This approach is designed to simultaneously optimize two key objectives: ensuring the helpfulness of the output and minimizing any potential harm, making the responses of LLMs more aligned with Human preferences. \citet{kwon2023reward} develop a proxy reward function using LLMs like GPT-3, which is created by first providing the LLM with a description of the behaviors desired by the user, along with a small number of examples. The LLM then produces rewards by evaluating how closely the outputs of a model align with the provided descriptions, essentially measuring their relevance to the established ground truth. \citet{scheurer2023training} propose Imitation Learning from Language Feedback, in which a language model is utilized to improve various outputs generated by a model. This refinement is based on a reference provided by a human. Following this process, the most effectively refined output is chosen to be used in further supervised fine-tuning. As outlined by \citet{kim-etal-2023-aligning}, ALMoST involves condensing human preferences into a set of heuristic guidelines. An example of such a rule is the idea that larger LLMs that utilize more comprehensive and higher-quality prompts are likely to yield superior responses. Based on these established guidelines, comparison data is generated using responses from LLMs of different sizes and with varying prompts. This data is then used to train a reward model. 
\citet{yang2023rlcd} propose Reinforcement Learning from Contrast Distillation, which aims to align language models without relying on human feedback. This approach involves training a preference model using simulated pairs of preferences, including both high-quality and low-quality examples which are generated through contrasting prompts, positive and negative. 

\citet{lee2023rlaif} further highlight the effectiveness of RLAIF. This work proposes that RLAIF not only matches but in some cases surpasses RLHF, and interestingly, RLAIF can also enhance the performance of Supervised Fine-Tuning. Another notable discovery is that directly prompting the LLM for reward scores during reinforcement learning can be more effective than the conventional approach of training a reward model based on LLM preferences. \citet{wang2023openchat} propose Conditioned-RLFT, which treats different data sources as coarse-grained reward labels and develops a class-conditioned policy to effectively utilize the varying qualities of data, which is a Reinforcement Learning-free supervised learning approach. \citet{cui2023ultrafeedback} propose a large-scale, high-quality, and diversified preference dataset labeled by GPT4 for comprehensive feedback. \citet{tunstall2023zephyr}, by proposing distilled Direct Preference Optimization \cite{rafailov2023direct} on UltraFeedback, obtaining a small by powerful LLM.

\subsubsection{Value}

Attaining alignment with human preferences allows large models to optimize human satisfaction by operating in a manner that aligns with human preferences. However, to establish trustworthy LLMs, the notion of 'aligning LLMs with human values' is proposed and the key principles of alignment are often summarized as the “HHH” criteria: helpful, harmless, honest~\cite{weidinger2021ethical, askell2021general}. Numerous methods have been undertaken for building trustworthy LLMs. However, due to the intrinsic difficulty of this aim, which is still an unsolved problem for proprietary models \cite{sun2024trustllm}, most existing methods rely on constructing high-quality human preference datasets~\cite{ji2023beavertails, solaiman2021process, bai2022training, qiu2022valuenet, kiesel-etal-2022-identifying, liu-etal-2022-aligning}, utilizing human-written rules as constrains~\cite{glaese2022improving, sun-etal-2023-moraldial, sun2023principledriven}, etc. For detailed progress on trustworthy LLMs, please further refer to \citet{yao2023instructions, liu2023trustworthy, sun2024trustllm}.

Though slightly under-explored, aligning LLMs with human values by distilling is still possible~\cite{bai2022constitutional, cui2023ultrafeedback, yang2023rlcd, sun2023principledriven}. For instance, \citet{bai2022constitutional} propose RLAIF, utilizing AI-generated labels to interactively improve both helpfulness and harmlessness. 
\citet{sun2023principledriven} prompt the student model with 16 principles as guidelines for generating helpful, ethical, and reliable responses.
Similarly, both harmless and harmful generations could be elicited by modifying the prompts, and then are used to train the preference model~\cite{yang2023rlcd}.
\citet{cui2023ultrafeedback} utilize GPT-4 to rank generations regarding helpfulness, truthfulness, and honesty.
\citet{liu2023training} advance the alignment of LLMs with societal values by incorporating simulated social interactions into the training process. This approach encompasses a range of elements, including demonstrations that are both in alignment and in conflict with social norms, as well as collective ratings, in-depth feedback, and responses that are revised iteratively.

\subsection{Agent}
\subsubsection{Tool Using}

While recent LLMs have shown proficiency in solving various tasks, they still tend to make mistakes when handling large numerical values or executing intricate mathematical calculations \cite{qian2022limitations, she2023pitfalls, manikandan2023language, liang2023taskmatrixai, mialon2023augmented}. Thus equipping LLM agents with the capability to utilize tools has been increasingly focused on. 
Commonly used methods mainly relied on human-curated data for training \cite{parisi2022talm, nakano2022webgpt, qin-etal-2023-webcpm, song2023restgpt} or prompt designing\cite{cai2023large, shen2023hugginggpt, hao2024toolkengpt}. 
Recently, distillation-based methods are also proposed \cite{schick2023toolformer, zhang2023graph, patil2023gorilla, tang2023toolalpaca, qin2023toolllm, yuan2023craft, gao2023confucius, wang2024mllmtool, shen2024small, yuan2024easytool}. 

Toolformer \cite{schick2023toolformer} utilizes a self-supervised manner, avoiding large human annotations, to obtain the most required APIs to use and further distill this capability to the model itself. The performance of the GPT-J-based Toolformer surpasses OPT (66B) \cite{zhang2022opt} and GPT3 (175B) \cite{brown2020language} greatly. 
Graph-ToolFormer \cite{zhang2023graph} aims to equip LLMs with the ability to process and reason over complex graph data, which is designed to enhance LLMs with graph reasoning skills using external graph reasoning API tools by adopting ChatGPT to annotate and augment a larger graph reasoning statement dataset for training. 
Gorilla \cite{patil2023gorilla} addresses the limitations of current LLMs in generating accurate input arguments and reduces the problem of "hallucination" or generating incorrect API usage and it collects thousands of models from platforms like HuggingFace and Torch Hub as the API calls and utilizes GPT4 to generate synthetic instruction data for training. 
GPT4Tools \cite{yang2023gpt4tools} introduces to enable open-source LLMs like LLaMA and OPT to use multimodal tools, a capability previously limited to advanced proprietary models like ChatGPT and GPT-4. The approach involves generating an instruction-following dataset by prompting an advanced teacher model with multimodal contexts, using the Low-Rank Adaptation optimization. 
ToolAlpaca \cite{tang2023toolalpaca} proposes a framework aimed at enhancing the tool-use capabilities of compact language models for embodied intelligence. It creates a dataset with 3938 instances from over 400 real-world tool APIs across 50 categories and utilizes ChatGPT to generate documentation for each prompt for later training. 
ToolLLM \cite{qin2023toolllm} proposes a comprehensive framework for enhancing LLMs with tool-use proficiency, focusing on data creation, model training, and evaluation by distilling from chatGPT. Their ToolLLaMA shows impressive performance in executing complex instructions and handling new APIs, rivaling ChatGPT.
CRAFT \cite{yuan2023craft} builds a general tool creation and retrieval framework, which utilizes GPT4 to generate code snippets as the created tools. During the inference, other small LLMs could select and retrieve from the generated code snippets to execute or generate other methods conditioned on the given snippets. 
Confucius \cite{gao2023confucius} introduces a tiered training strategy for LLMs to master tool usage through a graduated curriculum and an innovative method called Iterative Self-instruction from Introspective Feedback (ISIF) for dynamic dataset enhancement to handle complex tools. 
MLLM-Tool \cite{wang2024mllmtool} is a multi-modal tool agent capable of interpreting instructions embedded in visual or audio content through the integration of multi-modal encoders with open-source large language models. As a trainable method, the initial instruction-answer pairs are generated by utilizing GPT4. 
\citet{shen2024small} demonstrate that small LLMs are weak tool learners and proposes a multi-LLM framework that decomposes the tool-use ability of a single model into a planner, caller, and summarizer for the tool using, leading to a supreme performance. The two-stage training strategy introduced by this work is powered by ChatGPT and GPT4 for collecting execution trajectories for the training set. 
\citet{yuan2024easytool} notice the potential issue of the current lengthy tool documentation, which hinders LLMs from understanding how to utilize a tool, thus proposing EASYTOOL to purify the important information from extensive documentation. The ground truth summarization of the training documents is obtained by using ChatGPT.

\subsubsection{Planning}

Another important aspect for LLM agents is the ability to decompose high-level tasks to a chosen set of actionable steps \cite{huang2022language}, which is especially useful when acting in interactive environments. \citet{huang2022language} first demonstrate that LLMs can generate plausible goal-driven action plans without training, introduces non-invasive tools to enhance model executability, and assesses these methods through human evaluation to balance executability and semantic accuracy. Most existing methods utilize prompting strategies for task planning \cite{singh2022progprompt, zhou2023leasttomost, song2023llm, wang2023describe, yao2023tree, liu2023llm+, hao2023reasoning, hu2023tree}, or building human-curated data for training \cite{lin2023grounded, valmeekam2023on}. Recently, there have also been some distilling methods emerging \cite{chen2023fireact, zeng2023agenttuning, yin2023lumos, qiao2024autoact, kong2023tptuv2}. 

FireAct \cite{chen2023fireact} introduces an innovative approach for refining LLMs. This method involves fine-tuning smaller-scale LLMs using agent trajectories that are derived from a variety of tasks and prompting techniques. Applying this method with trajectories generated by GPT4 has been shown to consistently enhance performance.
AgentTuning \cite{zeng2023agenttuning} aims to enhance the performance of LLMs in executing agent tasks without sacrificing their wide-ranging capabilities. By utilizing a new dataset called AgentInstruct, which includes high-quality interaction trajectories, it applies a hybrid instruction-tuning approach that merges these trajectories with general domain instructions. 
Lumos \cite{yin2023lumos} pertains to a novel framework designed to train agents using a unified data format and modular architecture based on open-source LLMs. This system comprises three key modules: planning, grounding, and execution, enabling the decomposition of tasks into subgoals and actionable steps. 
TPTU-v2 \cite{kong2023tptuv2} focuses on improving the task planning and tool usage abilities of LLMs in real-world scenarios, by utilizing data generated by human experts or LLMs. It introduces a framework comprising three components: an API Retriever, an LLM Finetuner, and a Demo Selector. 
AUTOACT \cite{qiao2024autoact} proposes an agent learning framework that does not require large-scale annotated data or synthetic trajectories from high-resource models like GPT-4. Instead, it uses a self-instruct method to generate its own planning trajectories with limited initial data. It then applies a division-of-labor strategy, creating sub-agents specialized in different aspects of the task completion process. 

Distillation also works out for the training of embodied multi-modal agents~\cite{sumers2023distilling, yang2023embodied, ma2023eureka, pmlr-v202-du23f, sumers2023distilling}. 
For instance, \citet{sumers2023distilling} aim to enhance the ability of AI agents to follow instructions by using pretrained vision-language models to provide supervision for understanding and acting upon language within their operational environment, leveraging model distillation and hindsight experience replay to teach them contextually relevant interactions in a simulated 3D setting. 
Emma \cite{yang2023embodied} evaluates the challenges and inefficiency of training an embodied agent in a noisy visual world without expert guidance, 
and proposes to train them in a simulated environment using imitation learning, guided by an expert Language Model (like ChatGPT), which operates in a corresponding text-based simulation, focusing on the same tasks.

\subsection{NLP Task Specialization}

NLP tasks often grapple with challenges like data scarcity, interpretability issues, privacy concerns, and noisy data. 
The ``Knowledge" section of our survey illustrates various methods for distilling knowledge from LLMs, effectively setting the stage for student models to adapt to a range of NLP tasks.
This knowledge provides supervision for the training of student models through information augmentation (e.g., CoT and explanation), data augmentation, and semantic representation.
By transferring the distilled knowledge from LLMs, student models can better handle diverse NLP challenges, improving task performance and addressing data limitations more robustly.

\subsubsection{Natural Language Understanding}

Natural Language Understanding (NLU) is a fundamental NLP task that involves comprehending and interpreting human language. The knowledge distilled from LLMs, such as through data labeling or augmentation, is typically transferred into encoder-based language models like BERT~\cite{vaswani2017attention} and RoBERTa~\cite{liu2019roberta}. 

Regarding the task of classification, certain studies have been noteworthy~\cite{dai2023auggpt, Gilardi_2023, he-etal-2023-targeted, DBLP:conf/iclr/GaoPLXY0ZLLK23, chenglin2023mixed, li2023distilling}. AugGPT~\cite{dai2023auggpt} focuses on both general and clinical domain text classification. To address the limitations of small-scale clinical datasets, which often lack expert annotation and are subject to stringent privacy regulations, AugGPT utilizes knowledge from teacher LLMs to rephrase each sentence in the training samples. This process creates multiple conceptually similar but semantically distinct samples, enhancing the dataset's richness and diversity.
Another approach is demonstrated by \citet{Gilardi_2023}, who employ ChatGPT as an annotator to categorize inputs. This method has been shown to outperform crowd-workers in several tasks, including relevance, stance, topics, and frame detection.
Furthermore, \citet{he-etal-2023-targeted} propose \textit{Targeted Data Generation} (TDG), a novel approach for identifying challenging subgroups within a dataset. TDG leverages LLMs, along with human-in-the-loop, to generate new data specifically tailored for these subgroups, thereby enriching the dataset and improving model performance in sentiment analysis and natural language inference tasks.
To facilitate the clinical information extraction task, \citet{tang2023does} elicit diverse samples from LLMs by providing examples and different seeds of clinical entities, i.e. the \textit{Curation} manner.

Several studies have also focused on multiple NLU tasks~\cite{ding2023gpt3, he2023annollm, wang2021zerolabel, DBLP:journals/tacl/HeNKH022, ye2022zerogen, DBLP:conf/nips/MengHZH22}. For example, \citet{he2023annollm} utilize the knowledge in GPT-3.5 to annotate inputs with labels and explanations for various NLU tasks, including user input and keyword relevance assessment, BoolQ, and WiC.
\citet{wang2021zerolabel} employ few-shot prompts to expand high-quality training data using GPT-3, i.e. the \textit{Expansion} manner.
Beyond merely employing a single approach to elicit NLP task knowledge, \citet{ding2023gpt3} explore a combination of \textit{Labeling}, \textit{Expansion}, and \textit{Curation} methods to extract knowledge from GPT-3 for distilling data for both sequence- and token-level NLP tasks.

\subsubsection{Natural Language Generation}

Natural Language Generation (NLG) is a key aspect of evaluating the capabilities of LLMs, encompassing tasks such as summarization, machine translation, and other open-ended text generation tasks. Known for their potent generative abilities and creativity, LLMs excel in these areas, making them prime sources for distilling knowledge into student models tailored for NLG tasks~\cite{xu-etal-2023-inheritsumm, xu2023recomp, ramnath2023tailoring, agarwal2023gkd}.
Additionally, the knowledge distilled from LLMs can be effectively used for NLG task-specific data augmentation~\cite{jung2023impossible, wang-etal-2021-want-reduce, guo2023improving, yang2023neural, wang-etal-2023-umass, yang-etal-2023-data}. 
While the previous sections have focused on the works about open-ended generation and multi-turn dialogue, this part will specifically highlight the distillation techniques relevant to other NLG tasks.

Although automatic metrics often favor smaller, fine-tuned models in summarization tasks, human evaluators tend to prefer the summaries generated by LLMs. Addressing this discrepancy, \citet{xu-etal-2023-inheritsumm} develop a student summarization model by distilling a GPTSUMM dataset, which comprises over 4 million paragraph-summary pairs generated by querying GPT-3.5. In a different approach, \citet{jung2023impossible} introduce `Impossible Distillation,' a method that creates high-quality summarization-specific dataset from weak teacher LLMs. This method involves training a student model on the generated dataset and enhancing its capabilities through Self-Knowledge. 
Turning to the task of machine translation, where creating parallel corpora is traditionally expensive and time-consuming, \citet{yang2023neural} propose a three-step distillation process. This process involves generating seeds of verbs and nouns, forming sentences, and then translating these sentences. Their findings suggest that while the distilled dataset may lack diversity, it effectively improves the translation signal for training student translation models.
To distill high-quality content-grounded data automatically, Genie~\cite{yehudai2024genie} proposes a general methodology containing three key steps: (a) preparation of the content, (b) distillation of responses from a teacher LLM corresponding to the content, and (c) filtering mechanism to ensure the quality and faithfulness of the generated data. Genie demonstrates that student models trained through this distilled data can match or even surpass models trained on human-generated data.

\subsubsection{Information Retrieval}

Information Retrieval (IR) represents a crucial branch of computer science, focused on efficiently retrieving information relevant to user queries from extensive repositories~\cite{cai2022hyper,liu2022adam,feng2023knowledge,shen2023large}. A typical IR system encompasses three main components: the query rewriter, the retriever, and the reranker. Recent studies have highlighted the effectiveness of employing LLMs in IR systems, e.g. in enhancing the reranking stage through both point-wise and list-wise ranking methods~\cite{ma2023zeroshot, sun2023chatgpt, qin2023large}. However, the practical application of LLMs in IR systems faces challenges, primarily due to their slower generation speed, which conflicts with the low-latency requirements of IR tasks~\cite{sun2023chatgpt}. As a result, the KD of LLMs emerges as a more promising approach for IR, offering a way to infuse the distilled knowledge from LLMs into various stages of the IR pipeline without compromising on speed. There has been a significant body of work demonstrating how knowledge distilled from LLMs can benefit each component of the IR system, including the \textit{Query Rewriter}~\cite{srinivasan-etal-2022-quill, ma-etal-2023-query}, the \textit{Retriever}~\cite{DBLP:conf/iclr/DaiZMLNLBGHC23, sachan-etal-2022-improving, sachan-etal-2023-questions, schick-schutze-2021-generating, meng2023augtriever, peng2023soft}, and the \textit{Reranker}~\cite{DBLP:journals/corr/abs-2202-05144, sun2023chatgpt, pradeep2023rankvicuna, pradeep2023rankzephyr, DBLP:conf/emnlp/Saad-FalconKSFF23,  10.1145/3539618.3592067, jeronymo2023inparsv2, sun2023instruction}.

\vspace{2mm}
\pa{Query Rewriter.}
The Query Rewriter (QR) is a pivotal component in IR systems, tasked with enhancing the precision and expressiveness of user queries by refining or modifying the initial query to more accurately align with the user's information needs. 
One notable approach is QUILL~\cite{srinivasan-etal-2022-quill}, which introduces a two-stage distillation method for query intent understanding. Initially, a retrieval-augmented LLM, serving as the `professor,' is distilled into a non-retrieval augmented teacher LLM, aiming to bolster its understanding capabilities. Subsequently, this enhanced teacher LLM is distilled into a final student model using a large dataset, further refining the process.
Incorporating the QR into IR systems, \citet{ma-etal-2023-query} develop a 'Rewrite-Retrieve-Read' framework. This process begins with an LLM rewriting the queries via prompting, followed by a retrieval-augmented reading stage. To integrate the rewritten queries effectively into the IR system, the knowledge gleaned from the LLM is distilled into a compact student rewriter. This rewriter is then fine-tuned using feedback from the LLM reader through reinforcement learning.

\vspace{2mm}
\pa{Retriever and Reranker.}
In IR systems, the Retriever is designed to efficiently locate the top-k relevant texts from a large corpus. It encodes both queries and documents into vector representations and performs retrieval by computing the dot product between these vectors.
The Reranker further refines the order of the retrieved documents to improve the overall quality of the output. This is achieved in two primary ways, including \textit{Pointwise Reranker} and \textit{Listwise Reranker}. 
Pointwise Reranker takes both the query and a single candidate document as input to directly generate a relevance score.
Listwise Reranker directly reorders a list of input documents in terms of their relevance.

\textit{Retriever and Pointwise Reranker.} For the retriever and pointwise reranker, a common application of KD from LLMs is the generation of pseudo-queries for given documents. This approach aims to expand the pairwise data, enhancing the training of dense retrievers or rerankers.
For example, InPars~\cite{DBLP:journals/corr/abs-2202-05144} utilizes GPT-3 to generate multiple pseudo-queries for an unlabeled document. To ensure the relevance of these queries, the system filters them based on the highest log probabilities of generating a query conditioned on the documents. Subsequently, InPars fine-tunes a reranker based on monoT5~\cite{10.5555/3455716.3455856}. Another similar approach, Promptagator~\cite{DBLP:conf/iclr/DaiZMLNLBGHC23}, introduces a few-shot dense retrieval method that leverages a small number of demonstrations from the target domain for pseudo-query generation.
Diverging from the reliance on unlabeled documents, \citet{sachan-etal-2022-improving} distill knowledge from GPT-4 to curate diverse synthetic data for text embedding tasks across nearly 100 languages. They fine-tune powerful decoder-only LLMs, such as Mistral-7b~\cite{jiang2023mistral}, on this synthetic data using standard contrastive loss. Remarkably, this method demonstrates strong performance on text embedding and multilingual retrieval benchmarks without any labeled data.
Beyond generating pseudo-queries, teacher LLMs can also be employed to generate relevance scores as soft labels. These scores are used to train the retriever by minimizing the KL-divergence loss between the teacher and student distributions, as explored by \citet{sachan-etal-2023-questions}.

\textit{Listwise Reranker.} A distinct set of studies focuses on listwise reranking, where its advantage lies in comparing multiple documents simultaneously to determine the optimal reorder. RankGPT~\cite{sun2023chatgpt} leverages GPT-4 to generate permutations for a group of candidate passages. To distill this listwise ranking knowledge into a pointwise student reranker, various training loss functions are employed, such as Listwise Cross-Entropy~\cite{DBLP:conf/ictir/BruchWBN19}, RankNet~\cite{10.1145/1102351.1102363}, and LambdaLoss~\cite{10.1145/3269206.3271784}. Building upon RankGPT's framework, RankVicuna~\cite{pradeep2023rankvicuna} and RankZephyr~\cite{pradeep2023rankzephyr} further refine this approach by directly fine-tuning a listwise reranker using teacher-generated textual permutations. This enables the student reranker to produce sequences of ranked results directly, bypassing the intermediate step of calculating individual relevance scores.

\subsubsection{Recommendation}

Recommender systems are integral to enhancing user experience in various online services, providing personalized content based on user preferences and behaviors. 
Many works have demonstrated that LLMs could be directly used as recommenders without fine-tuning~\cite{wang2023generative, 10.1145/3604915.3610646} or generate auxiliary textual features to benefit recommender systems~\cite{xi2023openworld, ren2023representation, wei2024llmrec}.
~\cite{wang-etal-2023-llm4vis, ren2023representation, wei2024llmrec}. 
However, the real-time nature of online recommender systems demands rapid response times, posing a challenge with the inherent inference latency associated with LLMs. To address this, several studies have explored ways to distill and integrate the knowledge from LLMs into recommender systems, thereby leveraging their advanced capabilities while mitigating latency issues for efficient real-time recommendations~\cite{10.1145/3604915.3608829, zhang2023recommendation, liu2023once}.

\citet{10.1145/3604915.3608829} tackle data scarcity in narrative-driven recommendation (NDR), where users provide detailed descriptions of their preferences. They utilize GPT-3 to create synthetic narrative queries from user-item interactions via few-shot prompting, then distill this data into retrieval models for NDR. Similarly, GENRE~\cite{liu2023once} employs GPT-3.5 to augment datasets with new knowledge about news summarization, user profiles, and personalized content, aiding the training of content-based recommendation models. 
To bridge the gap between language models and recommender systems, some research views behavior modeling as an extension of language modeling~\cite{cui2022m6rec, liu2023pretrain}. InstructRec~\cite{zhang2023recommendation}, for instance, interprets recommendation as instruction following. They use ChatGPT to distill a wealth of user-personalized instruction data reflecting diverse preferences and intentions based on real historical interactions. This data is then used to fine-tune a 3B student language model specifically for recommendation purposes.

\subsubsection{Text Generation Evaluation}
Text generation evaluation, i.e. NLG evaluation, focuses on assessing the quality of generated content. Unlike traditional NLG evaluation metrics like BLEU~\cite{10.3115/1073083.1073135} or ROUGE~\cite{lin-2004-rouge}, which primarily rely on surface-level text comparisons, LLMs, trained on extensive corpora and refined through techniques like RLHF, offer a more human-aligned assessment. This sophistication has led to the increasing use of LLMs in NLG evaluation (detailed further in ~\cite{li2024leveraging}). Through KD of LLMs, student evaluators could enhance inference efficiency and achieve more flexible and highly customized evaluation~\cite{wang2023pandalm, kim2023prometheus, xu-etal-2023-instructscore, jiang2023tigerscore, li2023generative}.

PandaLM~\cite{wang2023pandalm} concentrates on a pairwise evaluator designed to compare two pieces of generated content. It utilizes a teacher LLM (GPT-3.5) to judge which response is better for a given instruction and input, providing reasons for its decision. Addressing the need for customized and flexible criteria to meet realistic user demands, Prometheus~\cite{kim2023prometheus} distills GPT-4 to construct a training dataset that includes reference answers and a variety of customized scoring rubrics. This dataset is then used to tune LLaMA for evaluating model-generated responses.
Instructscore~\cite{xu-etal-2023-instructscore} takes a more fine-grained approach by using GPT-4 to create detailed analysis data. This data is employed to tune LLaMA, enabling it to perform error analysis on generated texts compared to reference texts. The system further refines its evaluation capabilities through self-training with real model-generated response-reference pairs.
For reference-free evaluation across diverse domains, TigerScore~\cite{jiang2023tigerscore} samples data from a variety of text generation datasets, such as summarization, translation, and data-to-text. It distills error analysis knowledge from GPT-4 and uses this to fine-tune LLaMA.
Lastly, to adapt evaluation to real-world scenarios beyond conventional NLP tasks, Auto-J~\cite{li2023generative} collects real-world user queries and their evaluations from a teacher LLM. This massive dataset of real-world scenarios is then used to distill evaluation knowledge into LLaMA through fine-tuning, enhancing its practical applicability.

\subsubsection{Code}
LLMs, trained on extensive corpora containing code, are highlighted for their proficiency in code-related tasks. Their capabilities extend beyond direct code generation to include the provision of external knowledge and data, which is crucial in distilling their expertise into smaller, more efficient models. Several works have successfully distilled code knowledge from LLMs into those compact and specialized code models~\cite{codealpaca, rozière2023code, gunasekar2023textbooks, wei2023magicoder, chen2023personalised, liu2023mftcoder, yu2024wavecoder, DBLP:journals/corr/abs-2311-14904,DBLP:journals/corr/abs-2308-14731,DBLP:journals/corr/abs-2312-15692}.

A primary focus in these student code models is on code generation, a task of both common utility and practical significance. For instance, Code Alpaca~\cite{codealpaca} fine-tunes Llama using self-instruct with ChatGPT-distilled instructions specifically for code generation tasks. Similarly, Code Llama-instruct~\cite{rozière2023code} is fine-tuned via self-instruct, prompting Llama-2~\cite{touvron2023llama} with coding problems, and further refined with unit tests. 
Phi-1~\cite{gunasekar2023textbooks} aims to enhance the quality of distilled code data by extracting ``textbook quality" data from a teacher LLM, incorporating Python textbook and exercise data. Magicoder~\cite{wei2023magicoder} addresses potential biases in teacher LLMs by referencing a wealth of open-source code, yielding more diverse and grounded data for code generation. To consider the capability of the student model and leverage the feedback of the teacher, PERsD~\cite{chen2023personalised} introduces a Personalized Distillation method where the teacher LLM refines the student's generated code based on the execution feedback of the executor.

However, these models primarily target the code generation task, lacking generalizability across a broader range of code-related tasks. To address this issue, MFTCoder~\cite{liu2023mftcoder} utilizes self-instruct to distill diverse code data from teacher LLMs for various tasks, such as code completion and text-to-code generation, training a student model via multi-task learning. WaveCoder~\cite{yu2024wavecoder}, in contrast, creates a comprehensive instruction tuning dataset covering four universal code-related tasks distilled from GPT-3.5-turbo. WaveCoder first selects a diverse coreset of raw data using the KCenterGreedy~\cite{DBLP:conf/iclr/SenerS18} clustering method, then employs the teacher LLM for generating task definitions and outputs. The teacher model also plays a role in evaluating and filtering this data. Notably, WaveCoder demonstrates superior generalization across different code-related tasks compared to other open-source models.

\subsection{Multi-Modality}

Multimodal Large Language Models (MLLMs) surpass traditional language-only LLMs by understanding and processing information across multiple modalities, more closely mirroring human perception and enabling a broader range of real-world applications. There is a growing trend towards developing MLLMs that follow multimodal instructions, facilitating tasks with enhanced levels of interactivity. 
To address the scarcity of multimodal instruction-following data and to harness the commonsense and world knowledge embedded in teacher LLMs, numerous studies have focused on multimodal knowledge distillation from LLMs~\cite{liu2023visual, zhao2023svit, wang2023believe, chen2023shikra, park2023localized, pi-etal-2023-detgpt,zhao2023chatspot, liu2023mitigating, wu2023nextgpt, luo2023valley, jiang2023iluvui, li2023stablellava, xu2023pointllm}.

\vspace{2mm}
\pa{Vision-Language.}
In the vision-language domain, LLaVA~\cite{liu2023visual} pioneers the extension of the Self-Instruct approach from the language to the multimodal field. It translates images into textual descriptions, including captions and bounding boxes, and distills GPT-4 for generating new data in the context of seed examples. This approach creates a LLaVA-Instruct-150k dataset, which serves as the foundation for further developments like LLaVA-1.5~\cite{liu2023improved} and GPT4ROI~\cite{zhang2023gpt4roi}, enhancing the instruction-following capabilities of MLLMs.
To expand the dataset's scale, SVIT~\cite{zhao2023svit} introduces a 4.2 million image dataset, distilled from GPT-4 by leveraging manual image annotations. It employs a novel data recipe to select an informative, diverse, and balanced subset of training data.
LVIS-Instruct4V~\cite{wang2023believe} leverages GPT-4V~\cite{2023GPT4VisionSC}, a powerful large multimodal model, as a teacher to distill a more accurate and context-aware instruction-following dataset, focusing on fine-grained understanding.
Further advancements include integrating specific region referencing in image-based instruction following. For instance, Shikra~\cite{chen2023shikra} uses GPT-4 to distill referential question-answer pairs from the Flickr30K~\cite{plummer2015flickr30k} dataset, enhancing the understanding of referential regions within images. LSKD~\cite{park2023localized} introduces localized references to specific image regions, prompting the teacher LLM to generate commonsense inferences about these areas.
To enhance the visual instruction tuning pipeline with text-rich images, LLaVAR~\cite{zhang2023llavar} employs the text-only GPT-4 as a teacher, using recognized texts and image captions to generate 16K conversation pairs for text-rich images. The resultant student MLLM demonstrates enhanced interaction skills in content that combines both text and imagery.

\vspace{2mm}
\pa{Multiple Modalities.} 
To extend knowledge distillation of LLMs to encompass more modalities, such as audio and video, several innovative approaches have been introduced. These methods typically involve transforming these modalities into a textual format comprehensible to teacher LLMs, followed by the distillation of the teacher.
Macaw-LLM~\cite{lyu2023macaw} leverages GPT-4 to generate instruction-response pairs corresponding to the content of images or videos. MIMIC-IT~\cite{li2023mimicit} aims to broaden the scope to language, image, and video understanding, creating a substantial dataset with 2.8 million multimodal instruction-response pairs distilled from ChatGPT. ChatBridge~\cite{zhao2023chatbridge}, on the other hand, represents a novel approach in multimodal language modeling. It translates various non-textual modalities into text, combining fine-grained and global descriptions. This information is then used to distill responses from ChatGPT or GPT-4 through an in-context learning process, effectively bridging the gap between different modalities.

\vspace{2mm}
\pa{Others.}
Beyond distilling instruction-following data, several methods have emerged that concentrate on harnessing different aspects of knowledge from LLMs. For instance, EMMA~\cite{yang2023embodied} trains an MLLM to act as an embodied reflex agent within a visual environment. It achieves this by distilling GPT-4's skills in a parallel textual world, generating actions and providing reflective feedback. Silkie~\cite{li2023silkie} takes a unique approach by distilling preferences from GPT-4V, focusing on criteria like helpfulness and visual faithfulness. \citet{ha2023scaling} represent another innovative direction, where it generates, labels, and distills diverse robot-centric exploration experiences by LLMs into a multi-task visuo-linguo-motor policy.

\tikzstyle{my-box}=[
    rectangle,
    rounded corners,
    text opacity=1,
    minimum height=1.5em,
    minimum width=5em,
    inner sep=2pt,
    align=left,
    fill opacity=.5,
]
\tikzstyle{leaf}=[my-box, minimum height=1.5em,
    text=black, align=left,font=\scriptsize,
    inner xsep=2pt,
    inner ysep=4pt,
]
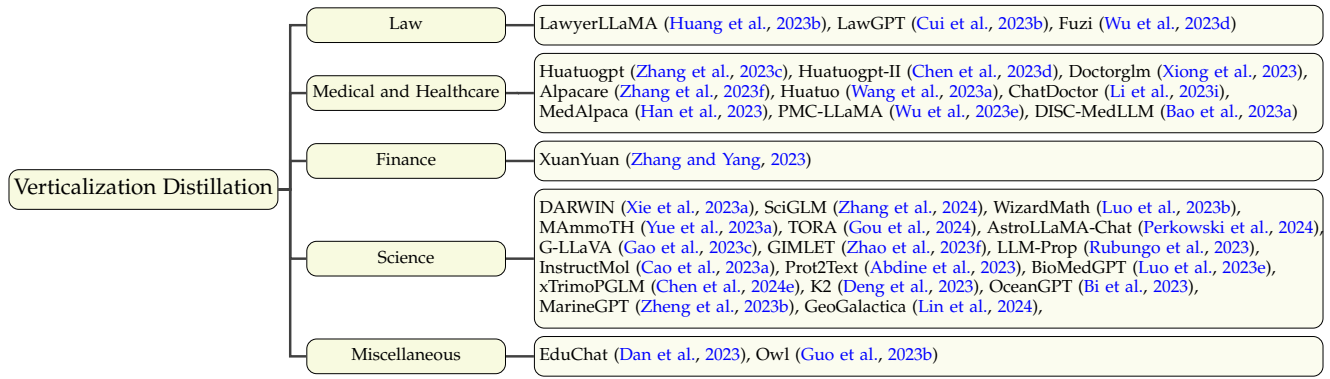
\begin{figure*}[ht!]
    \centering
    \resizebox{0.97\textwidth}{!}{
        \begin{forest}
            forked edges,
            for tree={
                grow=east,
                reversed=true,
                anchor=base west,
                parent anchor=east,
                child anchor=west,
                base=left,
                font=\small,
                rectangle,
                draw=black,
                rounded corners,
                align=left,
                text centered,
                minimum width=4em,
                edge+={darkgray, line width=1pt},
                s sep=3pt,
                inner xsep=2pt,
                inner ysep=3pt,
                ver/.style={rotate=90, child anchor=north, parent anchor=south, anchor=center},
            },
            where level=1{text width=8em,font=\scriptsize,}{},
            where level=2{text width=7em,font=\scriptsize,}{},
            where level=3{text width=9em,font=\scriptsize,}{},
            where level=4{text width=6.1em,font=\scriptsize,}{},
            [
                Verticalization Distillation, for tree={fill=hidden-yellow!50}
                [
                    Law
                    [
                        LawyerLLaMA~\cite{huang2023lawyer}{, }LawGPT~\cite{cui2023chatlaw}{, }Fuzi~\cite{sdu_fuzi_mingcha}, leaf, text width=33em
                    ]
                ]
                [
                    Medical and Healthcare
                    [
                        Huatuogpt~\cite{zhang2023huatuogpt}{, }Huatuogpt-II~\cite{Chen2023HuatuoGPTII}{, }Doctorglm~\cite{xiong2023doctorglm}{, }\\Alpacare~\cite{zhang2023alpacare}{, }Huatuo~\cite{wang2023huatuo}{, }ChatDoctor~\cite{li2023chatdoctor}{, }\\MedAlpaca~\cite{Han2023MedAlpaca}{, }PMC-LLaMA~\cite{wu2023pmc}{, }DISC-MedLLM~\cite{Bao2023DISC-MedLLM}, leaf, text width=33em                    
                    ]
                ]
                [
                    Finance
                    [
                        XuanYuan~\cite{Zhang2023xuanyuan}, leaf, text width=33em      
                    ]
                ]
                [
                    Science
                    [
                        DARWIN~\cite{Xie2023DARWIN}{, }SciGLM~\cite{Zhang2024SciGLM}{, }WizardMath~\cite{luo2023wizardmath}{, }\\MAmmoTH~\cite{yue2023mammoth}{, }TORA~\cite{Gou2023ToRA}{, }AstroLLaMA-Chat~\citep{Perkowski2024AstroLLaMAChat}{, }\\G-LLaVA~\citep{Gao2023GLLAVA}{, }GIMLET~\citep{Zhao2023GIMLET}{, }LLM-Prop~\citep{Rubungo2023LLM-Prop}{, }\\InstructMol~\citep{Cao2023InstructMol}{, }Prot2Text~\citep{Abdine2023Prot2Text}{, }BioMedGPT~\citep{luo2023biomedgpt}{, }\\xTrimoPGLM~\citep{xTrimoPGLM2024Chen}{, }K2~\citep{Deng2023K2}{, }OceanGPT~\citep{Bi2023OCEANGPT}{, }\\MarineGPT~\citep{Zheng2023MarineGPT}{, }GeoGalactica~\citep{Lin2024GeoGalactica}{, } , leaf, text width=33em      
                    ]
                ]
                [
                    Miscellaneous
                    [
                        EduChat~\citep{Dan2023EduChat}{, }Owl~\citep{Guo2023OWL}, leaf, text width=33em      
                    ]
                ]
            ]
        \end{forest}
    }
    \caption{Taxonomy of Verticalization Distillation.}
    \label{fig:vertical}
\end{figure*}

\section{Domain-specified Vertical Distillation}\label{sec:vertical}
This section shifts from skill distillation to examine KD of LLMs in various vertical domains, including Law, Medical \& Healthcare, Finance, and Science, etc. It delves into customizing distilled LLMs for these fields, showing its significant role in enhancing domain-specific AI applications. The taxonomy of these works is shown in Figure~\ref{fig:vertical}.

\subsection{Law}

Law holds a crucial position in molding societies, overseeing human interactions, and ensuring justice prevails. Informed decision-making, legal interpretation, and the provision of legal advice by professionals hinge on precise and current information. Legal intelligent applications in different scenarios usually require combinations of multiple fundamental capabilities of legal text retrieval, understanding, reasoning and generating~\cite{zhang2023unleashing,sun2023short,lai2023large}.
To address challenges like legal terminology, subtle interpretations, and the constant evolution of legislation presents distinctive challenges that demand customized resolutions. To handle the above challenges, several studies have investigated the customization of LLMs for intelligent legal services~\cite{cui2023chatlaw,yue2023disc,huang2023lawyer,sdu_fuzi_mingcha}. This involves a continued pre-training process on extensive legal corpora, followed by fine-tuning with self-constructed instructions or augmented data using advanced LLMs.

\citet{huang2023lawyer} have unveiled a Chinese legal large model named LawyerLLaMA. The model undergoes an initial pre-training phase on an extensive legal corpus, systematically assimilating knowledge of the Chinese legal system. Subsequently, fine-tuning occurs through the analysis of objective questions from the Chinese National Judicial Examination~\cite{zhong2020jec} and the gathering of responses to legal consultations using ChatGPT. This process equips the model with the ability to apply legal knowledge to specific scenarios.
\citet{cui2023chatlaw} present LawGPT, built upon the foundation of OpenLLAMA. The model is trained using a construction process that incorporates real-world legal text, legal regulations, judicial interpretations, and actual legal consultation data. Additionally, the authors utilize the ChatGPT API for assisted construction, enabling the generation of supplementary data derived from the existing dataset.
\citet{sdu_fuzi_mingcha} have developed a large-scale Chinese legal model (named Fuzi) with ChatGLM as its foundation. 
This model undergoes training on an extensive Chinese legal corpus, which incorporates unsupervised judicial language data, including diverse judgment documents and legal regulations. Additionally, it undergoes supervised judicial fine-tuning with data encompassing legal QA and case retrieval. Fuzi's training also involves both general instruction fine-tuning datasets, such as Alpaca, and domain-specific instruction fine-tuning datasets from LawyerLLaMA~\cite{huang2023lawyer} and LawGPT~\cite{cui2023chatlaw}.

\subsection{Medical and Healthcare}

% The integration of LLMs carries substantial promise in fundamentally reshaping the landscape of medical data analysis, comprehension, and smart medical services~\citep{Singhal2023MedPaLM2,Yang2023PLLaMa}. Significant research endeavors have been dedicated to adapting general-purpose LLMs to the medical domain, given the ever-expanding wealth of information encompassing electronic health records, medical literature, and clinical data. 
% Especially in healthcare, LLMs are revolutionizing patient care, research, and administrative efficiency. They enhance diagnostic accuracy by analyzing patient data and medical literature, offering personalized recommendations, and identifying potential drug interactions. LLMs also streamline administrative tasks by automating patient documentation and processing insurance claims, reducing the burden on healthcare providers and improving patient experiences. Furthermore, they facilitate medical research by synthesizing vast amounts of data to uncover new insights into diseases and treatments (will be discussed later).

% These adaptations extend across a spectrum, ranging from refining the precision of medical diagnoses~\cite{wang2023cmb} and providing personalized treatment recommendations~\cite{zhu2023promptcblue} to automating routine administrative processes within healthcare settings.

The integration of LLMs holds great potential for transforming medicine and healthcare. Extensive research has focused on adapting general-purpose LLMs to the medical domain~\cite{Singhal2023MedPaLM2}, such as electronic health records, and healthcare applications like patient care~\cite{zhu2023promptcblue}.
Recent work has focused on enhancing medical instruction-following data with advanced teacher LLMs to better align with complex user instructions. Given the abundance of medical data, most studies combine real-world data with distilled instruction data from teacher LLMs~\cite{zhang2023huatuogpt, xiong2023doctorglm, zhang2023alpacare, wang2023huatuo, li2023chatdoctor, Han2023MedAlpaca, Wu2023PMCLLAMA, Bao2023DISC-MedLLM, Chen2023HuatuoGPTII}. 

While existing studies predominantly concentrate on training using dedicated medical dialogue datasets comprising medical textbooks~\cite{wu2023pmc}, biomedical papers~\cite{luo2023biomedgpt} medical knowledge-graphs~\cite{bao2023disc}, or authentic doctor-patient interactions~\cite{bao2023disc}, an expanding body of research is delving into the augmentation of medical instruction-following data with advanced LLMs to enhance the alignment with practical user instructions.
\citet{zhang2023huatuogpt} introduce HuatuoGPT specifically tailored for medical consultations. The model leverages both \textit{distilled data from ChatGPT} and \textit{real-world data from doctors} during the supervised fine-tuning stage.
In a parallel effort, \citet{xiong2023doctorglm} construct a dataset of medical dialogues in Chinese, employing ChatGPT's assistance. Their methodology encompassed various techniques to train DoctorGLM, an easily deployable LLM designed for tasks such as diagnoses, drug recommendations, and other medical advice.
\citet{zhang2023alpacare} fine-tune LLaMA-series models using 52k diverse, machine-generated, medical instruction-following data named MedInstruct-52k. This effort resulted in the development of AlpaCare, a model demonstrating robust medical proficiency and generalizability across both general and medical-specific domain free-form instruction evaluations.
In a different vein, \citet{wang2023huatuo} propose HuaTuo, a LLaMA-based model that undergoes supervised fine-tuning with generated QA instances. This refinement process enhances the model's possession of more reliable medical knowledge.
\citet{li2023chatdoctor} introduce ChatDoctor, which was first trained as a generic conversation model based on LLaMA. It utilized 52K instruction-following data from Stanford University’s Alpaca project~\cite{alpaca}. Subsequently, the conversation model underwent fine-tuning on a dataset of 100K patient-physician conversations collected from an online medical consultation website. This two-step training process underscores the model's adaptability to diverse conversational contexts, particularly those specific to patient-physician interactions.

Built upon existing datasets, MedAlpaca~\citep{Han2023MedAlpaca} proposes to reconstruct the data with GPT-3.5-Turbo, which is then used to fine-tune LLMs for effective medical applications. 
Furthermore, PMC-LLaMA~\citep{Wu2023PMCLLAMA} proposes a training framework (i.e., continual pre-training and domain-specific multi-task supervised fine-tuning) to adapt a general LLM to the medicine domain, where GPT-4 is leveraged to write synonymous sentences for data augmentation in the SFT.
To adapt LLMs to real-world medical consultation, DISC-MedLLM~\citep{Bao2023DISC-MedLLM} leverages GPT-3.5 to 1) construct 50K QA pairs in a few-shot manner and 2) re-generate the 420k dialogues based on real cases, which are then used to train LLMs in a supervised fine-tuning manner. 
More recently, HuatuoGPT-II~\citep{Chen2023HuatuoGPTII} proposes a one-stage training with instruction-formatting unification of domain data collection for medical adaption upon LLMs, where GPT-4 is used to formulate medical questions to fine-tuning instructions.

These diverse studies collectively contribute to the advancing field of the medical domain, facilitated by knowledge distillation from advanced LLMs. Through the exploration of various methodologies, these approaches provide valuable insights into the challenges and potential breakthroughs at the intersection of cutting-edge language models and medical applications.

\subsection{Finance}

The application of LLMs to the finance domain \citep{Xue2023WeaverBird} significantly transforms how financial data is analyzed, decisions are made, and customer interactions are managed. In finance, LLMs offer unprecedented capabilities in understanding complex financial documents, predicting market trends, and automating risk assessment, thus enabling more informed and faster decision-making processes. By processing and analyzing vast amounts of unstructured financial data, such as news articles, reports, and real-time market feeds, LLMs can identify patterns and insights that were previously inaccessible, leading to more accurate forecasts and strategic financial planning. Furthermore, LLMs enhance customer experiences through personalized financial advice, automated customer service, and sophisticated chatbots that can handle complex queries. This level of automation and insight has the potential to increase efficiency, reduce operational costs, and improve compliance and risk management practices in financial institutions, making LLMs a transformative force in the finance sector.
Knowledge distillation from a proprietary LLM is still under-explored, and most existing works focus on adapting LLMs to finance applications by continual pre-training on finance-specific corpora \citep{Wu2023BloombergGPT,Lu2023BBTFin} or fine-tuning in a supervised manner on multi-task finance-specific instructions \citep{Yang2023InvestLM,Xie2023PIXIU,Wang2023FinGPT}.

Specifically, XuanYuan~\citep{Zhang2023xuanyuan} leverages self-instruct over seed data and self-QA over structured/unstructured data to generate instruction data in the finance domain, which is used to train a finance LLM.

\subsection{Science}

The integration of LLMs into the science domain \citep{Taylor2022Galactica,Yin2023FORGE} represents a paradigm shift in research, knowledge discovery, and the dissemination of scientific information. In science, LLMs are leveraged to digest and synthesize vast amounts of literature, aiding in the identification of new research opportunities and the acceleration of scientific breakthroughs. They facilitate the understanding of complex scientific concepts by summarizing research papers, generating hypotheses, and even drafting research proposals and manuscripts, thus significantly reducing the time researchers spend on literature review and enabling them to focus more on experimental work. LLMs also democratize access to scientific knowledge by providing layperson summaries of complex research findings, making science more accessible to non-experts and fostering a broader public understanding of scientific advancements. By enhancing the efficiency of research workflows and fostering interdisciplinary collaborations, LLMs are poised to accelerate the pace of scientific discovery and innovation across various fields.
To distill knowledge from an LLM, DARWIN Series~\citep{Xie2023DARWIN} utilizes a semi self-instruct for instruction generation for science papers, which is then used to fine-tune an LLM. 
SciGLM~\citep{Zhang2024SciGLM} proposes to train a scientific LLM, which prompts a teacher LLM to generate detailed answers for unlabelled scientific questions, as well as a self-reflective critic-and-revise to improve data quality.
Besides the above knowledge distillation methods to adapt LLMs to science, we will also delve into how the distillation happens in sub-domains, e.g., mathematics, astronautics, chemistry, etc.

\vspace{2mm}
\pa{Mathematics.} The application of LLMs within the sub-domain of mathematics heralds a transformative era in mathematical research, education, and problem-solving \citep{Azerbayev2023LLEMMA,Yu2023Outcome}. LLMs in mathematics facilitate the exploration and understanding of complex mathematical theories and problems by providing intuitive explanations, proofs, and solutions that can bridge the gap between advanced mathematical concepts and learners at various levels. 
These models have shown potential in conjecturing new mathematical theorems and patterns, thus opening new avenues for research and discovery that might not have been readily accessible to humans alone. In education, they serve as personalized tutors, offering students step-by-step guidance through mathematical problems and adapting explanations to the learner's level of understanding. This democratizes access to high-quality mathematical education and fosters a deeper appreciation and understanding of mathematics among a broader audience. By enhancing collaborative efforts through the generation of new ideas and the simplification of complex concepts, LLMs are poised to significantly advance the field of mathematics, making it more accessible, efficient, and innovative. 
WizardMath~\citep{luo2023wizardmath} enhances the mathematical reasoning capabilities of Llama-2 by applying the novel Reinforcement Learning from Evol-Instruct Feedback (RLEIF) method, significantly outperforming other open-source LLMs on the GSM8k and MATH benchmarks, as well as surpassing several closed-source LLMs including ChatGPT-3.5 and Minerva.
MAmmoTH~\citep{yue2023mammoth} is a series of open-source LLMS specifically developed for general math problem-solving, achieving superior performance on nine mathematical reasoning datasets. Utilizing a novel instruction tuning dataset called MathInstruct, which combines chain-of-thought and program-of-thought rationales, MAmmoTH models demonstrate substantial improvements over existing models.
TORA~\citep{Gou2023ToRA}, a series of Tool-integrated Reasoning Agents, significantly advances mathematical problem-solving by combining natural language reasoning with the use of external computational tools. It markedly outperforms existing open-source models on 10 mathematical reasoning datasets, showcasing notable improvements over both rationale-based and program-based approaches, and introduces innovative training techniques such as output space shaping to enhance model reasoning capabilities. 
G-LLaVA~\citep{Gao2023GLLAVA} introduces a significant advancement in geometric problem-solving for LLMs by leveraging a multimodal approach that combines text and image data. This model, utilizing the Geo170K dataset comprising over 170,000 geometric image-caption and question-answer pairs, demonstrates remarkable improvements over GPT-4V on the MathVista benchmark.

\vspace{2mm}
\pa{Astronautics.} The application of LLMs in astronautics \citep{Nguyen2023AstroLLaMA} propels the field forward.
AstroLLaMA-Chat~\citep{Perkowski2024AstroLLaMAChat} is an advancement of the AstroLLaMA model, leveraging a 7B-parameter LLaMA-2 model and targeted continual pre-training on a curated astronomy corpus to enhance performance in astronomy-focused question-answering. This model demonstrates significant improvements in specialized topic comprehension and introduces a chat-enabled version for the astronomy community, highlighting the effectiveness of domain-specific knowledge distillation in achieving superior performance on specialized topics.

\vspace{2mm}
\pa{Chemistry and Materials Science.} The integration of LLMs into Chemistry and Materials Science has revolutionized the way researchers approach the discovery and development of new compounds and materials. 
By analyzing vast datasets and scientific literature, LLMs can predict the properties and behaviors of substances, significantly accelerating the innovation cycle. 

GIMLET~\citep{Zhao2023GIMLET}, Graph Instruction based MolecuLe zEro-shoT learning, is a novel approach to molecule property prediction that integrates graph and text data within a single language model framework, aiming to improve instruction-based zero-shot learning for molecular tasks. By leveraging a transformer mechanism with generalized position embedding and decoupled attention, GIMLET significantly outperforms traditional molecule-text baselines in zero-shot learning scenarios, demonstrating the model's effectiveness in generalizing from instructions to a broad range of molecule-related tasks without prior explicit task-specific training.
LLM-Prop~\citep{Rubungo2023LLM-Prop}, leveraging the T5 model, showcases how LLMs can outperform SoTA graph neural networks in predicting the physical and electronic properties of crystalline solids from text descriptions. This approach underscores the potential of text-based methods in materials science, offering significant improvements in prediction accuracy while also contributing a benchmark dataset, TextEdge, to foster further research in this emerging field.
InstructMol~\citep{Cao2023InstructMol} integrates multi-modal data, aligning molecular structures with natural language instructions for drug discovery tasks. Through a novel two-stage instruction-tuning approach, it significantly enhances performance in molecule-related tasks, establishing a reliable molecular assistant that outperforms existing LLMs and reduces the performance gap with specialized models. This demonstrates the value of multi-modal integration in developing versatile tools for complex domains like drug discovery.

\vspace{2mm}
\pa{Biology.} 
In the field of Biology, particularly in the study of proteins, DNA, and RNA, LLMs are revolutionizing our understanding of the fundamental molecules of life. By analyzing vast datasets of biological sequences and structures, LLMs can predict the three-dimensional shapes of proteins, potential functions, and interactions at a scale and speed beyond traditional computational methods. 
This capability is critical for unraveling the complexities of biological systems, advancing drug discovery by identifying targets and designing molecules with high precision, and understanding genetic diseases through the interpretation of genomic variations. 

Prot2Text~\citep{Abdine2023Prot2Text} introduces a novel multimodal framework for generating protein function descriptions in free text by combining GNNs and LLMs. This approach, which integrates structural and sequential protein information, highlights the transformative impact of knowledge distillation through the fusion of GNNs and LLMs for accurate protein function prediction, potentially revolutionizing research in bioinformatics and biological sciences.
BioMedGPT~\citep{luo2023biomedgpt} introduces a multimodal generative pre-trained transformer specifically designed for the biomedicine domain, emphasizing the significance of aligning molecular, protein, and natural language modalities to enhance biomedical question-answering, molecule, and protein QA tasks. This framework showcases the critical role of knowledge distillation in bridging the gap between complex biological data and human language, thereby facilitating groundbreaking advancements in drug discovery and therapeutic target identification.
xTrimoPGLM~\citep{xTrimoPGLM2024Chen}, a unified 100B-scale pre-trained transformer model, addresses both protein understanding and generation tasks by integrating autoencoding and autoregressive pre-training objectives. Its significant advancements over existing models in 18 protein understanding benchmarks and its capability in de novo protein sequence generation highlight the model's importance in advancing the field of protein science through knowledge distillation.

\vspace{2mm}
\pa{Geography, Geology, and Environmental Science.}
The integration of LLMs into Geography, Geology, and Environmental Science is revolutionizing these fields by enhancing data analysis, predictive modeling, and interdisciplinary research \cite{roberts2023gpt4geo, lin2023geogalactica, wang2023nearrealtime}. 

K2~\citep{Deng2023K2}, the first-ever LLM specialized in the geoscience domain, demonstrates the significant impact of knowledge distillation in vertical domain specialization. By adapting the general-domain LLaMA-7B model with a 5.5B token geoscience corpus and introducing the GeoSignal instruction tuning dataset, K2 showcases enhanced performance in geoscience knowledge understanding and utilization. The model's development highlights a novel approach to efficiently gather domain-specific data and align model responses to specialized user queries.
OceanGPT~\citep{Bi2023OCEANGPT}, introduced as the first LLM for ocean science tasks, underscores the vital role of knowledge distillation in the vertical domain of oceanography. It leverages DOINSTRUCT, a novel framework for generating domain-specific instruction data through multi-agent collaboration, and establishes OCEANBENCH, a benchmark for evaluating LLMs in the ocean domain.
MarineGPT~\citep{Zheng2023MarineGPT} showcases the transformative potential of knowledge distillation in the marine domain by leveraging a novel vision-language model tailored for marine science. Utilizing the Marine-5M dataset, which includes over 5 million marine image-text pairs, MarineGPT excels in providing detailed, accurate, and domain-specific responses. 
GeoGalactica~\citep{Lin2024GeoGalactica} represents a pioneering step in specializing LLMs for geoscience, leveraging a 30 billion parameter model pre-trained on a vast geoscience corpus. This model is notable for being the largest of its kind within the geoscience domain.

\subsection{Miscellaneous}

Knowledge distillation of LLMs has vast potential across various verticals beyond the ones previously discussed, highlighting their versatility and transformative impact across different industries.
For instance, in the education sector, EduChat~\cite{Dan2023EduChat} exemplifies a chatbot system that provides tailored support to teachers, students, and parents. KD is central to its design, leveraging pre-training on educational data followed by fine-tuning with custom instructions to deliver capabilities such as essay evaluation and emotional support.
Similarly, Owl~\cite{Guo2023OWL}, an LLM designed for IT operations, boosts operational efficiency using the Owl-Instruct dataset, which is distilled from ChatGPT. By applying a mixture-of-adapter strategy for domain-specific tuning, it enhances analysis and performance in IT-related tasks.

\section{Open Problems}\label{sec:open}

\vspace{2mm}
\pa{Further Data Selection}
How much data is required for LLM distillation and how to filter out the low-quality data remain open-domain questions. In the field of instruction tuning, one of the most commonly used methods for distillation, \citet{zhou2023lima} propose that only $1000$ human-curated high-quality data is enough for the alignment of LLMs, hypothesizing that LLMs have learned the required knowledge from pretraining and only a small amount of data is required for the alignment. Its finding further raises a new question, how to automatically select the data for better distillation? 
\citet{chen2023alpagasus} directly apply ChatGPT to rate each data sample together with explanations, and then the data is selected based on the rating. 
\citet{cao2023instruction} split the existing instruction-tuning datasets and trains a linear function to select the most effective data based on their statistical properties. 
\citet{cherry} propose a data selection pipeline similar to self-distillation, in which the LLM firstly learns from a small subset of the data to get the basic ability, and then further uses this learned model to rate for the original dataset. 
\citet{du2023mods} propose to consider three aspects including quality, coverage, and necessity for the filtering process. 
\citet{li2023shot} select instruction data by evaluating their one-shot improvement on a hold-out set. 
\citet{Li2024SuperfilteringWD} recently propose Superfiltering, which is able to utilize small language models like GPT2 to filter out the high-quality subset from a given high-quality dataset. 
Despite the emergence of these works working on data filtering, How to efficiently select the optimal distillation data for LLMs, and How much data is required for distillation are still unsolved.

\vspace{2mm}
\pa{Reduce the Distillation Cost (Lightweight Methods)}
Despite the remarkable abilities of the latest LLMs, their significant resource requirements underscore the urgent need to find efficient solutions to overcome these challenges. Common ways to further reduce the distillation cost include Model Compression and Efficient Fine-Tuning. In the realm of Model Compression, Quantization \cite{frantar2023optimal, dettmers2022gptint, kim2023finequant, tao-etal-2022-compression, yao2022zeroquant, xiao2023smoothquant}, Parameter Pruning \cite{ma2023llmpruner, zhang2023loraprune, frantar2023sparsegpt}, and Low-Rank Approximation \cite{xu2023tensorgpt, li2023losparse} are commonly utilized. In the realm of Efficient Fine-Tuning, Parameter Efficient Fine-Tuning \cite{hu2023llmadapters, liu2022fewshot, wang-etal-2022-adamix, hu2021lora, li-liang-2021-prefix, liu-etal-2022-p}, and Memory Efficient Fine-Tuning \cite{dettmers2023qlora, kim2023memoryefficient, malladi2024finetuning} are utilized. A detailed survey on Efficient Large Language Models can be found here in \citet{wan2024efficient}. The problem that remains is how can we further compress the model and build effective distillation algorithms.

\vspace{2mm}
\pa{Multi-Teacher Distillation}
Most of the existing distilled models are distilled from a single teacher model, however, it is widely accepted that models trained with different sources of data have various capabilities. Thus a question arises: Is it possible to distill knowledge from different teacher models into one student model? BabyLlama \cite{timiryasov2023baby} proposes to distill the knowledge from both the GPT2 and LLaMA into the small-size student models. Ensemble-Instruct \cite{lee-etal-2023-ensemble} tries to generate both instructions and responses ensembled from several different LLMs with RougeL as the indicator. FUSELLM \cite{wan2024knowledge} externalizes the collective knowledge and unique strengths by leveraging the generative distributions of different LLMs aiming to train a student model beyond those of any individual source LLM. Despite the recent progress in this topic, it still remains an under-explored topic. 

\vspace{2mm}
\pa{Explore Richer Knowledge from Teacher LLMs}
As indicated in Table~\ref{tab:my_label}, the majority of teacher LLMs are closed-source due to their advanced capabilities. Consequently, current methodologies primarily focus on using the generations from these models as hard labels, training student models through simple supervised fine-tuning. However, beyond the straightforward imitation of output behaviors via hard labels, there is a growing interest in harnessing richer knowledge from teacher LLMs, including feedback and feature knowledge, as well as exploring diverse combinations of knowledge elicitation methods.
As highlighted in the \textit{Feedback} section, teachers can provide various types of feedback based on the student's outputs~\cite{lee2023rlaif, jiang2023lion, chen2023personalised}. Similarly, the \textit{Feature} section discusses how knowledge based on features, such as logits serving as soft labels, can offer deeper, intrinsic insights into the teacher model~\cite{gu2023knowledge, agarwal2023gkd}.
These explorations have demonstrated promising outcomes, suggesting that access to a broader spectrum of knowledge can significantly enhance student model performance beyond what is achievable through simple SFT distillation alone. This highlights the critical need for further research into varied knowledge extraction methods from teacher LLMs to augment the effectiveness of KD processes.

\vspace{2mm}
\pa{Overcoming Catastrophic Forgetting During Distillation} %[Chongyang]
Previous research has delved into the fine-tuning of LLMs to acquire the ability to follow instructions or transfer knowledge for forthcoming tasks, skills, or domains, leveraging advancements in LLM technology. Nevertheless, investigations have revealed that the continual fine-tuning of LLMs on particular datasets (skills, domains) can lead to a phenomenon known as catastrophic forgetting, wherein previously acquired knowledge and problem-solving abilities for earlier tasks are compromised~\cite{chen2023lifelong,kotha2023understanding,koloski2023measuring,wu2024continual,luo2023empirical}. 
Earlier studies in  machine learning and deep learning
have investigated various techniques to help mitigate forgetting during the fine-tuning or continue learning process, such as rehearsal, which entails periodically revisiting and training on past data~\cite{kirkpatrick2017overcoming,rostami2019complementary,rolnick2019experience}, as well as regularization methods like elastic weight consolidation~\cite{lee2017overcoming}, or dynamic architecture methods~\cite{mallya2018piggyback,wang2022learning,hu2023dense,chen2023lifelong}.
To address the challenges of catastrophic forgetting and to enhance the diversity of generated instructions in knowledge distillation for LLMs, \citet{jiang2023lion} randomly sample an instruction from the easy instructions and also prompt the generator to generate a new instruction that belongs to the same domain as the sampled one.
In a similar vein, \citet{li2023unlock} study the problem of instruction-tuning in multi-modal LLMs knowledge distillation and introduce a competitive distillation framework. The model tries to produce new instructions that differ in content but are similar in difficulty to the original pictures in the multi-modal augmentation phase, so as to alleviate catastrophic forgetting of the model and enhance the diversity of the instruction tuning pool.
\citet{chen2023lifelong} propose the
Lifelong-MoE (Mixture-of Experts) architecture based on general language models, which dynamically adds model capacity via adding experts with regularized pretraining. Additionally, 
the model also introduces implicit regularization via distillation of the knowledge from old experts and gatings to effectively preserve old knowledge.
\citet{zeng2023continual} propose a new generative-based rehearsal method as Dirichlet Continual Learning (DCL). This method combines task distribution modeling and knowledge distillation to mitigate catastrophic forgetting without requiring access to the old data. 
To evaluate the effectiveness of instruction tuning in the context of continuous learning tasks, \citet{zhang2023citb} introduce a more challenging yet practical problem called Continual Instruction Tuning (CIT) and also establish
a benchmark suite consisting of learning and
evaluation protocols.
Although current research has explored some simple methods to alleviate knowledge forgetting during model fine-tuning or knowledge distillation processes, effectively avoiding catastrophic forgetting across domains and skills remains a challenging issue. How to retain the original model's capabilities effectively during knowledge distillation or transfer processes is still a challenging problem.

\vspace{2mm}
\pa{Trustworthy Knowledge Distillation}
Trustworthiness in LLMs is paramount, encompassing attributes such as truthfulness, safety, fairness, robustness, privacy, and adherence to machine ethics~\cite{sun2024trustllm}. The rapid advancement of LLMs brings to the forefront concerns regarding their trustworthiness, stemming from their complex outputs, the biases present in vast training datasets, and the potential inclusion of private information.
Current efforts in KD of LLMs primarily focus on distilling various skills from LLMs, with relatively little attention paid to trustworthiness aspects. Existing studies tend to concentrate on a subset of trustworthiness aspects, such as helpfulness, honesty, and harmlessness~\cite{bai2022constitutional, yang2023rlcd, cui2023ultrafeedback}. Consequently, in the distillation process, student models may inherit issues related to trustworthiness from their teacher LLMs.
As assessed in \citet{sun2024trustllm}, smaller open-source LLMs generally fall short of their proprietary counterparts in trustworthiness metrics. Therefore, considering trustworthiness alongside the distillation of capabilities into student models is crucial. It is imperative that future research on KD not only enhances the capabilities of student models but also ensures that broader aspects of trustworthiness are meticulously addressed.

\vspace{2mm}
\pa{Weak-to-strong Distillation.} 
The concept of ``weak-to-strong generalization'' in LLMs \citep{Burns2024Weak} emphasizes the potential to leverage weak supervision to elicit the advanced capabilities of more powerful models. This approach challenges the traditional distillation paradigm by suggesting that even with limited or imperfect supervision, it is possible to enhance the performance of LLMs significantly. This necessitates exploring innovative strategies that enable weaker models to guide the learning process of stronger ones effectively, highlighting the importance of developing methods that can bridge the gap between these models. Such research could unlock new avenues for improving LLMs' efficiency and effectiveness, making the pursuit of ``weak-to-strong distillation'' a crucial area for future investigations in this LLM era. 
Initially, \citet{Burns2024Weak} investigates whether weak model supervision can unlock the full capabilities of much stronger models. Through experiments with pre-trained language models in the GPT-4 family across NLP, chess, and reward modeling tasks, it finds that finetuning strong models on weak labels leads to better performance than their weak supervisors, demonstrating weak-to-strong generalization. 
Then, \citet{Li2024Superfiltering} introduce Superfiltering, a method that employs smaller, weaker models like GPT-2 to select high-quality data for fine-tuning larger, more capable models such as LLaMA2. This approach is rooted in discovering a strong consistency in evaluating instruction tuning data difficulty across models of varying sizes. 
More recently, \citet{Ji2024Aligner} introduce Aligner, a novel approach for aligning LLMs with human values and intentions by utilizing weak supervisory signals from smaller models to improve the performance of larger models. 
However, \citet{Burns2024Weak} find that achieving the full capabilities of strong models requires more than naive finetuning, suggesting the need for further research in this area. Therefore, open questions still remain about 
1) What are the theoretical and practical limits of weak-to-strong distillation? Can weak supervision reliably extract and enhance the full spectrum of capabilities in stronger models across all domains, or are there inherent limitations based on model architecture or task specificity?
2) How do we identify or design the optimal weak supervisors for distilling knowledge into stronger models? Is there a framework or criteria to predict which weak models would be most effective in guiding the learning process of more complex models for specific tasks?
3) To what extent are weak-to-strong distillation techniques transferable and scalable across different sizes and types of models? How can these methods be adapted to ensure efficacy and efficiency in distilling knowledge from very large models to significantly smaller ones, especially in resource-constrained environments?

\vspace{2mm}
\pa{Self-Alignment.} 
Aligning LLMs traditionally relies heavily on human or teacher LLMs to supply extensive preference data. Consequently, the alignment of the student model is limited by the quantity of distilled preference data and the teacher's capabilities. Self-alignment offers a promising alternative, aiming to enhance alignment beyond the constraints of teacher-provided preferences. In self-alignment, the student model endeavors to autonomously improve and align its responses with desired behaviors, including generating model-written feedback, critiques, and explanations.
Several studies have explored utilizing the student model's inherent capabilities to generate knowledge for alignment~\cite{bai2022constitutional, sun2023principledriven, li2023selfalignment, yuan2024selfrewarding}. Beyond merely producing improved responses~\cite{bai2022constitutional, sun2023principledriven}, implementations of self-alignment include employing the student as its reward model to offer feedback~\cite{yuan2024selfrewarding}, a strategy that merges \textit{Self-Knowledge} with \textit{Feedback} methods of eliciting knowledge. We advocate for increasingly leveraging the student model itself to provide feedback, thereby enhancing self-alignment capabilities. This approach not only facilitates moving beyond traditional human/teacher preference-based rewards but also opens avenues for continual self-improvement and alignment.

\section{Conclusion and Discussion}\label{sec:conclude}

This survey has explored the diverse landscape of knowledge distillation for LLMs, highlighting key techniques, applications, and challenges. KD plays a crucial role in democratizing access to advanced LLM capabilities, providing cutting-edge advancements without the high costs of training and deployment. Our review emphasizes various KD approaches, from algorithmic innovations to skill enhancement and vertical distillation. Notably, data augmentation and synthesis within KD emerge as vital tools for improving distillation, revealing the powerful synergy between enriched training data and effective model distillation.
As the AI landscape evolves, rapid advancements in model architectures and training methods present both challenges and research opportunities for KD of LLMs. Future innovation will need to focus on achieving efficiency, transparency, and ethics while maintaining model trustworthiness. Furthermore, promising areas such as weak-to-strong generalization, self-alignment, and multi-modal LLMs offer the potential to enhance the capabilities of distilled models.
In conclusion, the KD of LLMs is set to play a pivotal role in the future of AI research. As highlighted in this survey, sustained research efforts will be critical in developing accessible, efficient, and responsible AI for all. 
Importantly, when conducting KD of LLMs like ChatGPT or Llama, it's essential to comply with the model providers' terms\footnote{OpenAI Business Terms:  \url{https://openai.com/policies/business-terms}}, such as the restrictions on developing competitive products.

\bibliographystyle{IEEEtranN}
\bibliography{custom}

\end{document}